\documentclass{article}

\usepackage{microtype}
\usepackage{graphicx}
\usepackage{listings}
\usepackage{subfigure, float}
\usepackage{booktabs} 
\usepackage{tabularx}
\usepackage{multirow}
\usepackage{array}
\usepackage{paralist, tabularx}
\usepackage{soul}
\usepackage[ruled]{algorithm2e}
\SetKwComment{Comment}{/* }{ */}

\usepackage{hyperref}

\usepackage[accepted]{icml2023}

\usepackage{amsmath}
\usepackage{amssymb}
\usepackage{mathtools}
\usepackage{amsthm}
\usepackage{makecell}
\usepackage{bm}
\usepackage{bbm}
\usepackage[capitalize,noabbrev]{cleveref}

\theoremstyle{plain}

\theoremstyle{definition}

\theoremstyle{remark}

\usepackage[textsize=tiny]{todonotes}

\usepackage{xspace}

\icmltitlerunning{SceneCraft: An LLM Agent for Synthesizing 3D Scene as Blender Code}

\begin{document}

\twocolumn[

\icmltitle{SceneCraft: An LLM Agent for Synthesizing 3D Scene as Blender Code}


\icmlsetsymbol{equal}{*}

\begin{icmlauthorlist}
\icmlauthor{\ \ \ Ziniu Hu}{caltech,google}
\icmlauthor{\ \ \ Ahmet Iscen}{google}
\icmlauthor{\ \ Aashi Jain}{google}
\icmlauthor{\ \ Thomas Kipf}{gdm}
\icmlauthor{\ \ \ Yisong Yue}{caltech}
\icmlauthor{\ \ David A. Ross}{google}
\icmlauthor{\ \ Cordelia Schmid}{google}
\icmlauthor{\ \ Alireza Fathi}{google}
\end{icmlauthorlist}

\icmlaffiliation{caltech}{California Institute of
Technology}
\icmlaffiliation{google}{Google Research}
\icmlaffiliation{gdm}{Google DeepMind}

\icmlcorrespondingauthor{Ziniu Hu}{ziniu@google.com}
\icmlcorrespondingauthor{Yisong Yue}{yyue@caltech.edu}
\icmlcorrespondingauthor{Alireza Fathi}{alirezafathi@google.com}

\icmlkeywords{Machine Learning, ICML}

\printAffiliationsAndNotice{}

\vskip 0.3in
]

\footnote{\textsuperscript{*}Correspondence to: Ziniu Hu $<$ziniu@google.com$>$, Yisong Yue $<$yyue@caltech.edu$>$ and
Alireza Fathi $<$alirezafathi@google.com$>$.}



\begin{abstract}
This paper introduces SceneCraft, a Large Language Model (LLM) Agent converting text descriptions into Blender-executable Python scripts which render complex scenes with up to a hundred 3D assets.
This process requires complex spatial planning and arrangement. We tackle these challenges through a combination of advanced abstraction, strategic planning, and library learning. 
SceneCraft first models a scene graph as a blueprint, detailing the spatial relationships among assets in the scene.
SceneCraft then writes Python scripts based on this graph, translating relationships into numerical constraints for asset layout. 
Next, SceneCraft leverages the perceptual strengths of vision-language foundation models like GPT-V to analyze rendered images and iteratively refine the scene. 
On top of this process, SceneCraft features a library learning mechanism that compiles common script functions into a reusable library, facilitating continuous self-improvement without expensive LLM parameter tuning.
Our evaluation demonstrates that SceneCraft surpasses existing LLM-based agents in rendering complex scenes, as shown by its adherence to constraints and favorable human assessments. 
We also showcase the broader application potential of SceneCraft by reconstructing detailed 3D scenes from the Sintel movie and guiding a video generative model with generated scenes as intermediary control signal.

\end{abstract}

\begin{figure}[t!]
    \centering
    \includegraphics[width=0.51\textwidth]{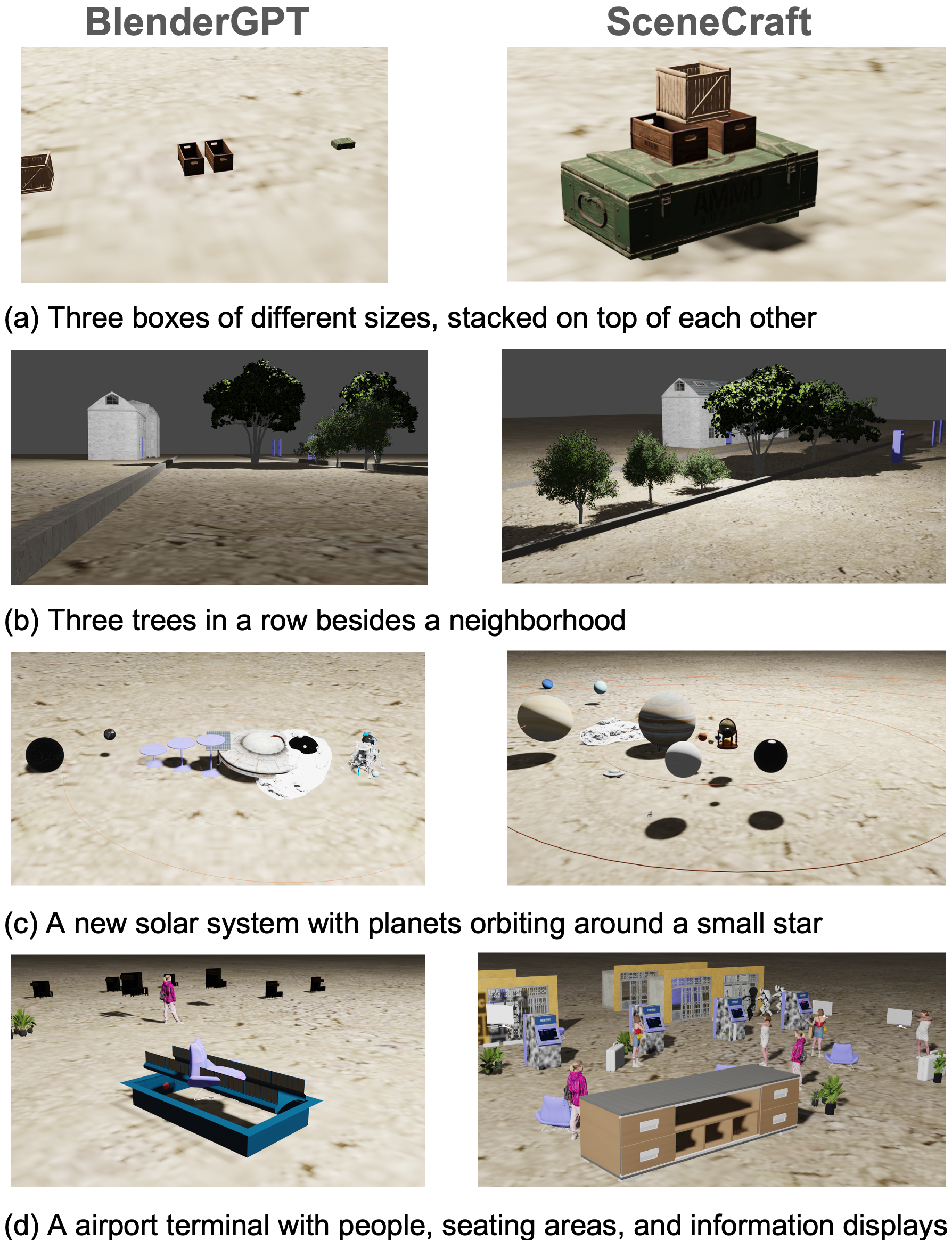}
    \vspace{-0.2in}
    \caption{
    Examples comparing SceneCraft's output against a BlenderGPT baseline for different queries.}
    \label{fig:teasing}
    \vspace{-0.2in}
\end{figure}

\section{Introduction}
Transforming natural language descriptions into 3D scenes is a key technology for industries like architectural design, game development, virtual reality, and cinematic production. Recent 3D generative models like DreamFusion~\citep{DBLP:journals/corr/abs-2209-14988} and Magic3D~\citep{DBLP:conf/cvpr/Lin0TTZHKF0L23} have made great progress in transforming text to a 3D neural representation of an object.
However, these works fall short of composing entire scenes with multiple assets due to dataset scale limitations and domain specificity.
In this work, we are inspired by how human artists typically adopt a holistic process for designing 3D scenes, where they take an iterative, step-by-step approach that includes storyboarding, 3D modeling, texturing, rigging, layout,  animation, and rendering, using professional software such as Blender\footnote{\url{https://www.blender.org/}}.
This iterative process grants the artists in studios a nuanced control over each asset’s placement and movement — a level of control not yet achieved by existing models. 

Our paper introduces SceneCraft, an LLM-powered agent that is designed to streamline this text-to-3D scene conversion process, closely emulating the workflow of studio artists. SceneCraft transforms textual descriptions into executable Blender code, rendering 3D scenes that are visually cohesive and contextually accurate.
This task goes beyond mere data processing, demanding a nuanced understanding of spatial and semantic relationships, which remains a challenge even for today's LLMs. While earlier systems like WordsEye~\cite{DBLP:conf/siggraph/CoyneS01} and SceneSere~\citep{DBLP:journals/corr/ChangESM17} have made progress towards using predefined templates and rules to extract spatial constraints from linguistic queries, they depend on extensive human input, especially in new domains with unique relationship patterns. 

\begin{figure}[t!]
  \centering
  \includegraphics[width=0.51\textwidth]{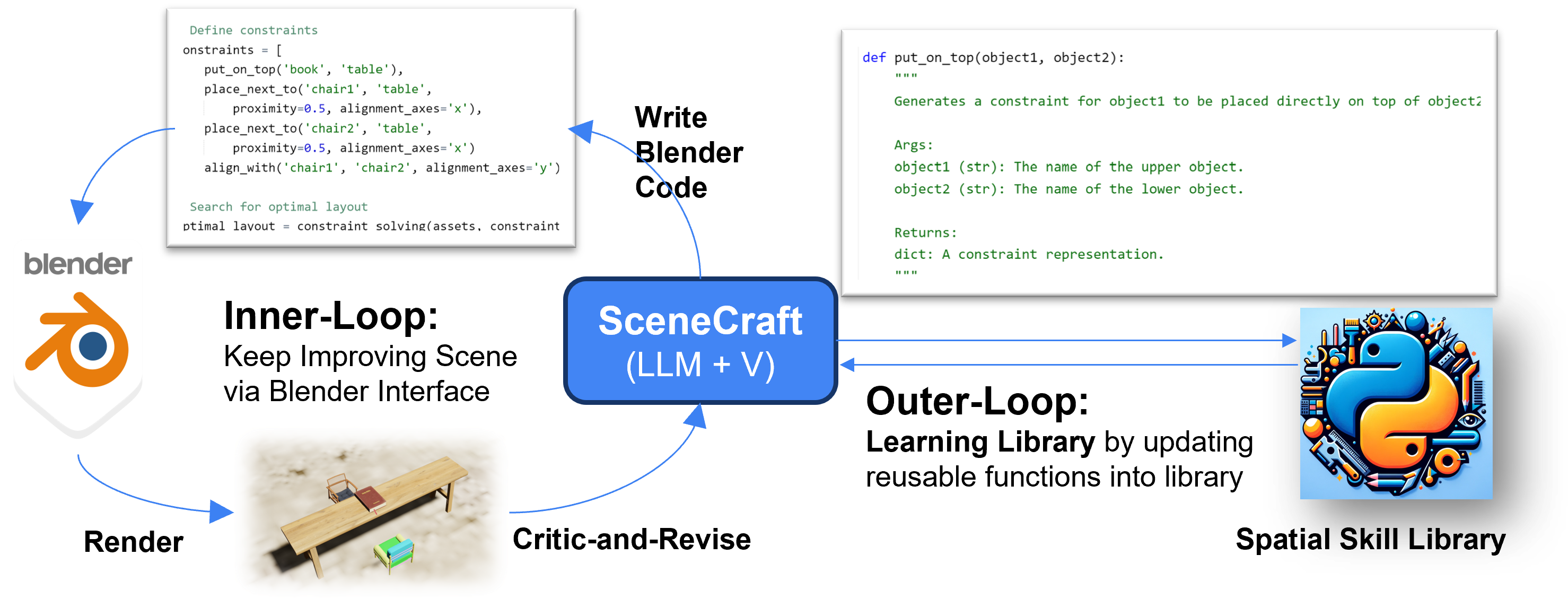}
  \vspace{-0.1in}
  \caption{SceneCraft is composed of a \textit{dual-loop} self-improving pipeline: in the inner-loop, per each scene, an LLM autonomously writes a script to interact with Blender, receives rendered image, and keeps improving the script until getting good scenes;
  in the outer-loop, SceneCraft summarizes common functions over a batch of written scripts to maintain a reusable design skill library.}
  \label{fig:overview}
  \vspace{-0.2in}
\end{figure}




SceneCraft leverages LLMs to autonomously generate Python code, translating spatial relations within scenes into precise numerical constraints. 
To achieve this, the core of SceneCraft is a dual-loop optimization pipeline, illustrated in Figure~\ref{fig:overview}. 
The inner-loop focuses on per-scene layout optimization. 
Here, an LLM-based planner constructs a scene graph outlining the spatial constraints for asset arrangement. SceneCraft then writes Python code to transform these relations into numerical constraints. These constraints are fed to a specialized solver that determines the layout parameters of each asset, including location, orientation and sizes. 
After rendering these scripts into images via Blender, we utilize a multimodal LLM (GPT-V~\citep{openai-gpt4vision}) to assess the alignment between the generated image and the textual description. If a misalignment is detected, the LLM identifies the problematic semantic relations and corresponding constraints, subsequently refining the scripts. This iterative process of refinement and feedback is crucial for enhancing the scene’s fidelity, ensuring each rendition progressively aligns more closely with the original vision, which also matches more the human artists' designing process.

Following the inner-loop refinement of scene scripts, SceneCraft starts its outer loop to dynamically expand its 'spatial skill' library. Within this procedure, it reviews the incremental changes made to the constraint scripts across inner-loop iterations for each scene, identifying and integrating common code patterns, thereby streamlining the acquisition of new non-parametric skills for self-improvement.  
For instance, if the text query describes a lamp placed on a desk, but the initial rendering shows desk lamps floating mid-air, SceneCraft may learn to introduce a new "grounded" constraint between lamp and desk surfaces. 
By continuously updating its library through such outer-loop learning over batches, SceneCraft acquires an expanding repertoire of spatial skills over time. SceneCraft is therefore able to handle increasingly complex scenes and descriptions without external human expertise or LLM parameter tuning. 

To evaluate SceneCraft, we conduct comprehensive experiments on both synthetic and real-world datasets. First, we create our own curated datasets with ground-truth spatial constraints to quantify SceneCraft's fidelity in translating text to constraint scripts. Second, we apply SceneCraft to the Sintel movie dataset by finetuning a video generative model on the first half of the movie conditioned on ground-truth scene images. For the second half, we generate scenes using SceneCraft and other baselines as input to the video model. Across datasets, results demonstrate SceneCraft's superior sample efficiency and accuracy in rendering intricate 3D scenes from textual descriptions, enabled by its dual-loop optimization. Quantitatively, SceneCraft achieves over 45.1\% and 40.9\% improvement on generated scenes' CLIP score, compared with another popular LLM agent baseline BlenderGPT, over both unseen synthetic queries and real-world movies like Sintel. SceneCraft also achieves significantly better constraint passing score (88.9 against 5.6).
Qualitatively, the scenes and videos generated using SceneCraft more accurately encapsulate the narrative and artistic nuances described in the text. It receives much higher human preference ratings on different perspective, and also benefit a video generative model with very light fine-tuning.
Together, our comprehensive evaluation validates SceneCraft as an adaptable and efficient framework for translating imaginative text to 3D reality while continuously improving itself.



This paper's contributions are:
\begin{compactitem}
    \item  An LLM Agent that transforms an input text query into a 3D scene by generating a Blender script. The script is iteratively improved by a multimodal LLM that identifies unsatisfied constraints and fixes them in a feedback loop. 

    \item A spatial skill library learned given a set of synthetic input queries without requiring human involvement and LLM fine-tuning, resulting in improved scene generation results.

    \item 
    Experimental results show that comparing with BlenderGPT, another LLM-based agent baseline, SceneCraft achieves 45.1\% and 40.9\% improvement on generated scenes' CLIP score, over both unseen synthetic queries and real-world movies like Sintel.  
\end{compactitem}

\begin{figure*}
    \centering
    \includegraphics[width=1\textwidth]{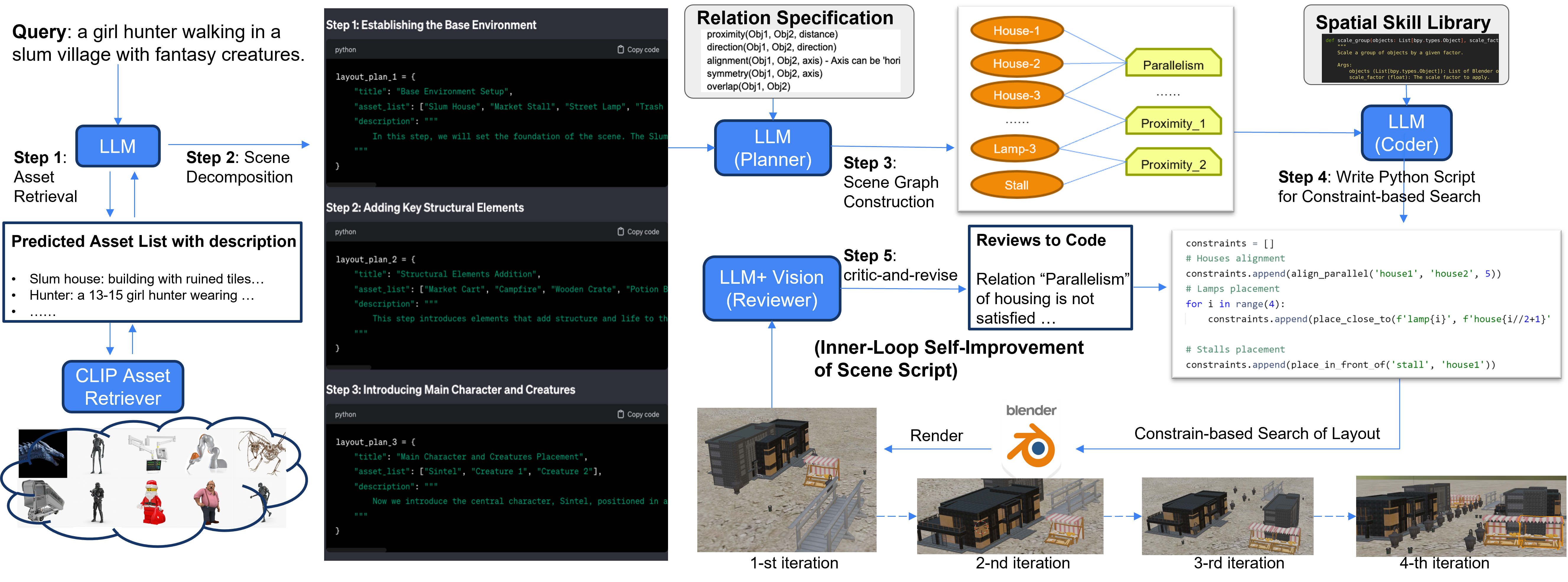}
    \vspace{-0.3in}
    \caption{\textbf{The workflow of SceneCraft's inner-loop improvement of each scene}. 1) given query, a LLM writes a list of assets descriptions, then use CLIP retriever to fetch assets; 2) then LLM decomposes the full query into a sequence of sub-scene, each associated with a subset of assets and a text description; 3) a LLM-Planner generate a relational graph linking assets to spatial relationship; 4) Based on the graph, LLM-Coder writes python codes to get a list of numerical constraints, which can be executed to search optimal layout, and render into image using Blender; 5) LLM-Reviewer with vision perception capability criticize the rendered image, and update the script accordingly. This critic-and-revise procedure can be done multiple times to iteratively improve the script and scene. }
    \label{fig:inner-loop}
\end{figure*}

\section{Approach}
Our goal is to transform a text query $q$ into a 3D scene $s$
that is not only spatially coherent but also contextually rich and aesthetically pleasing. This requires (a) identifying the correct spatial and contextual relationships between assets, and (b) predicting a high fidelity and nice looking arrangement that aligns with these relationships.

SceneCraft performs this task by building on top of a state-of-the-art multimodal LLM (i.e., GPT-4V~\citep{openai-gpt4vision}) and a professional rendering software (Blender). We now describe the key components of our method.

\subsection{Asset Retrieval and Scene Decomposition}
A scene consists of a set of assets, where each asset $a_i$ is a 3D model. Given the input text query $q$, the agent makes an LLM call to generate a list of asset names and description that shall be put in the scene. Based on them, a set of 3D assets ${\mathcal{A}}$ are retrieved from a large repository of 3D objects utilizing a CLIP-based retriever. The retrieval process first finds the top-10 assets based on the text description of each asset. Then each retrieved asset is rendered as an image and the one with the highest text-to-image score is selected.

Some scenes might contain up to a hundred assets, making the layout planning very difficult. Therefore, SceneCraft agent decomposes the scene into a set of sub-scenes, each representing a part of the entire scene. Breaking the problem into small pieces is a widely adopted strategy in natural language question answering~\citep{DBLP:conf/emnlp/PerezLYCK20} and general reasoning~\citep{DBLP:conf/iclr/ZhouSHWS0SCBLC23}.

The agent calls a LLM-empowered decomposer that breaks the input query into a sequence of sub-scenes $\hat{s}_k$, each containing a title, a list of asset names $\mathcal{A}_k$ and a sub-scene description $q_k$. The scene descriptions are used to guide the scene optimization in the later stages. 
\begin{align}
  (q_1, \mathcal{A}_1), \ldots, (q_K, \mathcal{A}_K)\gets \texttt{LLM-decomposer}(q).
\end{align}
As an example shown in Figure~\ref{fig:inner-loop}, given a query "a girl hunter walking in a slum village with fantasy creatures", SceneCraft decomposes it into three different steps, among which the first step includes the following information:
\begin{figure}[H]
    \centering
    \includegraphics[width=0.5\textwidth]{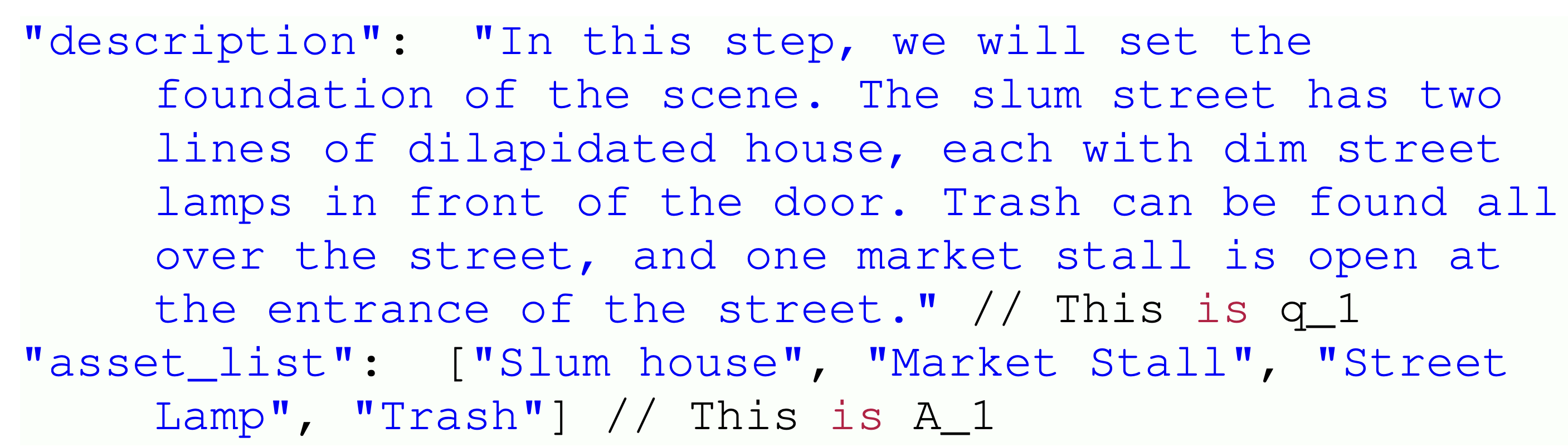}
\end{figure}

\subsection{Scene Graph Construction}
In order to put the 3D assets together to create a scene $s$, each asset $a_i$ requires its corresponding layout matrix $\mathcal{L}(a_i)$, encapsulating the position, scale and orientation of $a_i$ in the scene's coordinate frame. For instance, in a scene described as ``a round table with a vase on it, placed near a window'', the assets $a_i$, $a_j$ and $a_k$ could represent the `table', `vase' and the `window', respectively. Their layout matrices $\mathcal{L}(a_i)$, $\mathcal{L}(a_j)$ and $\mathcal{L}(a_k)$ shall position the vase, table and window in the scene such that the vase is standing on the table and the table is located near the window.

The key challenge is to correctly put each asset in the right location and orientation by predicting the layout matrix $\mathcal{L}(a_i)$ for each asset. The naive approach is to directly predict all the layout matrices directly given the scene description. However, this is a highly complex task even for the most advanced LLMs, due to the vast combinatorial space of the possible layouts and deep understanding the intricate spatial relations between assets.
This is why in SceneCraft we use a relational scene graph as an intermediate layer of abstraction.

To model the spatial relations between assets, SceneCraft utilizes a set of spatial and contextual relations, such as proximity, alignment, parallelism, etc. Each relation $r$ applies to a specific set of assets within the scene. 
Full list of relations that we consider can be found in Sec~\ref{sec:relation} in Appendix.

Using these relations, the scene $s$ is abstracted into a relational bipartite graph \( G(s) = (\mathcal{A}, \mathcal{R}, \mathcal{E}) \), which contains two types of nodes:
 \( \mathcal{A} \) represents the set of assets and $\mathcal{R}$ represents the set of relations as nodes. $\mathcal{E}$ represents the edges connecting a relation node to a subset of assets $\mathcal{E}(r)$ in the scene that satisfies this relation.\footnote{For each relation type, we can have multiple relation nodes linking to different subsets of assets, e.g., \textit{Align-1} relation node links table and vase, and \textit{Align-2} links vase and window.}

Based on this definition, SceneCraft then uses a \texttt{LLM-Planner} to construct a scene graph connecting assets to corresponding spatial relation nodes. 
\begin{align}
   \mathcal{G}(s) = (\mathcal{A}, \mathcal{R}, \mathcal{E}) \gets \texttt{LLM-Planner}(q_k, \mathcal{A})
\end{align}
For example, when we create the outline of slum village, $\texttt{LLM-Planner}$ predicts the following edges:
\begin{compactitem}
\item $\langle$Alignment, list of houses$\rangle$: all housess are aligned side by side to form a side-street;  
\item $\langle$Parallelism, two list of houses$\rangle$: Duplicate one side of street to form a pathway or road;  
\item $\langle$Proximity, each lamp, each house$\rangle$: lamps are located in front of each house.
\end{compactitem}




The relations between the assets provide soft spatial constraints for the layout matrices $\mathcal{L}$ of the assets. 
Thus, this intermediate graph serves as a high-level plan for subsequent code generation and self-improvement, which significantly reduces the complexity of arranging the assets in the scene.

\subsection{Scene Layout Optimization in a Feedback Loop}

After we obtain the spatial constraints between the assets, we use a set of scoring functions (one per relation) to optimize the scene layout. In Sec. \ref{sec:library_learning} we describe how we learn the library of scoring functions automatically. The scoring function $F_r(\cdot)$ for  relation $r$ captures whether the constraint is satisfied via
$F_r\big(\{\mathcal{L}(a_i) \mid a_i \in \mathcal{E}(r)\big\}, \texttt{arg}_r)$. $F_r$ takes as input a list of layout matrices $\mathcal{L}(a_i)$ connected by $r$, as well as the function arguments $\texttt{arg}_r$ such as the distance, directions, etc. $F_r$ returns a real number between 0 and 1 describing how much this relational constraint is satisfied. 

An \texttt{LLM-Coder} then reuses these existing functions stored in skill library to synthesize an overall Blender code script $\texttt{code}$, including loading the assets, doing grouping and generating all the numerical constraints, etc. 
The \texttt{LLM-Coder} will also predict all the arguments $\texttt{arg}_r$ for each function $F_r$, such as the exact distance for Proximity relation, etc.
\begin{align}
   \texttt{code}, \texttt{arg} \gets \texttt{LLM-Coder}(\mathcal{G}(s), q_k)
\end{align}
For each scene $s$ abstracted by a relational scene graph $G(s)$ finding an optimal layout could be formalized as the following optimization problem:
\begin{equation}
\hat{\mathcal{L}} \gets \underset{\mathcal{L}}{\mathrm{argmax}} \sum_{r \in \mathcal{R}} F_r\Big( \big\{\mathcal{L}(a_i) \mid a_i \in \mathcal{E}(r)\big\}, \texttt{arg}_r\Big) \label{eq:solve}
\end{equation}
This enables SceneCraft to simultaneously balance multiple constraints, ensuring a comprehensive and contextually accurate scene layout planning. After getting the optimal layout $\hat{\mathcal{L}}$, we can render the scene with the Blender code script $\texttt{code}$ to get image output. Examples of generated scripts and rendered images can be found at Sec.~\ref{sec:example} in Appendix.


\paragraph{Self-Improvement of Scene Script}
However, the agent often does not produce the correct layout outright. This is because either (a) the predicted constraints do not reflect the requirements in the input query or do not follow common-sense knowledge, which requires updating the scene graph edges $\mathcal{E}$ and the Blender code $\texttt{code}$); or (b) the generated constraint functions do not correctly reflect the semantic relationships and result in an incorrect layout (we need to update scoring functions $\mathcal{F}_r$ and arguments $\texttt{arg}_r$).

We iteratively improve the initially generated scene layout in a visual feedback loop by taking advantage of the perception capabilities of a multimodal LLM (GPT-V~\citep{openai-gpt4vision}). We render the generated scene into an image, then feed the rendered image and the scene description directly to the \texttt{LLM+V-Reviewer}, asking it which constraints are lacking or not correctly satisfied, asking it to revise the script to reflect all the mistakes it finds. 
If \texttt{LLM+V-Reviewer} finds out that the error is rooted in the constraint functions, it can either modify existing functions or add new sub-functions to improve the layout planning for the current scene. This procedure repeats at every iteration of the feedback loop. 
We denote that at the $t$-th iteration, the function for each relation is $\mathcal{F} ^{t}$ and the graph edges are $\mathcal{E} ^{t}$, and the predicted optimal layout is $\hat{\mathcal{L}^{t}}$. This shares a similar intuition with recent works that utilize foundational models to generate a reward signal~\citep{DBLP:journals/corr/abs-2312-09187, DBLP:journals/corr/abs-2310-12921, DBLP:journals/corr/abs-2310-12931, shinn2023reflexion}.
The feedback-loop optimization procedure can formally be written as:
\begin{align}
&\mathcal{E}^{(t+1)}, \mathcal{F}^{(t+1)}, \texttt{arg}^{(t+1)} \gets  \texttt{LLM+V-Reviewer}(\texttt{img} , q_k) \nonumber  \\
& \text{subject to} \ \  \texttt{img} \gets \texttt{Blender-Render}(\mathcal{A}, \mathcal{L}^{t}, \texttt{code}^{t})  \nonumber  \\
& \mathcal{L}^t \gets \underset{\mathcal{L}}{\mathrm{argmax}} \sum_{r \in \mathcal{R}} F^t_r\big(\big\{\mathcal{L}(a_i) \mid a_i \in \mathcal{E}^t(r)\big\}, \texttt{arg}^t_r\big) 
\end{align}
\begin{figure}[t!]
    \centering
    \includegraphics[width=0.53\textwidth]{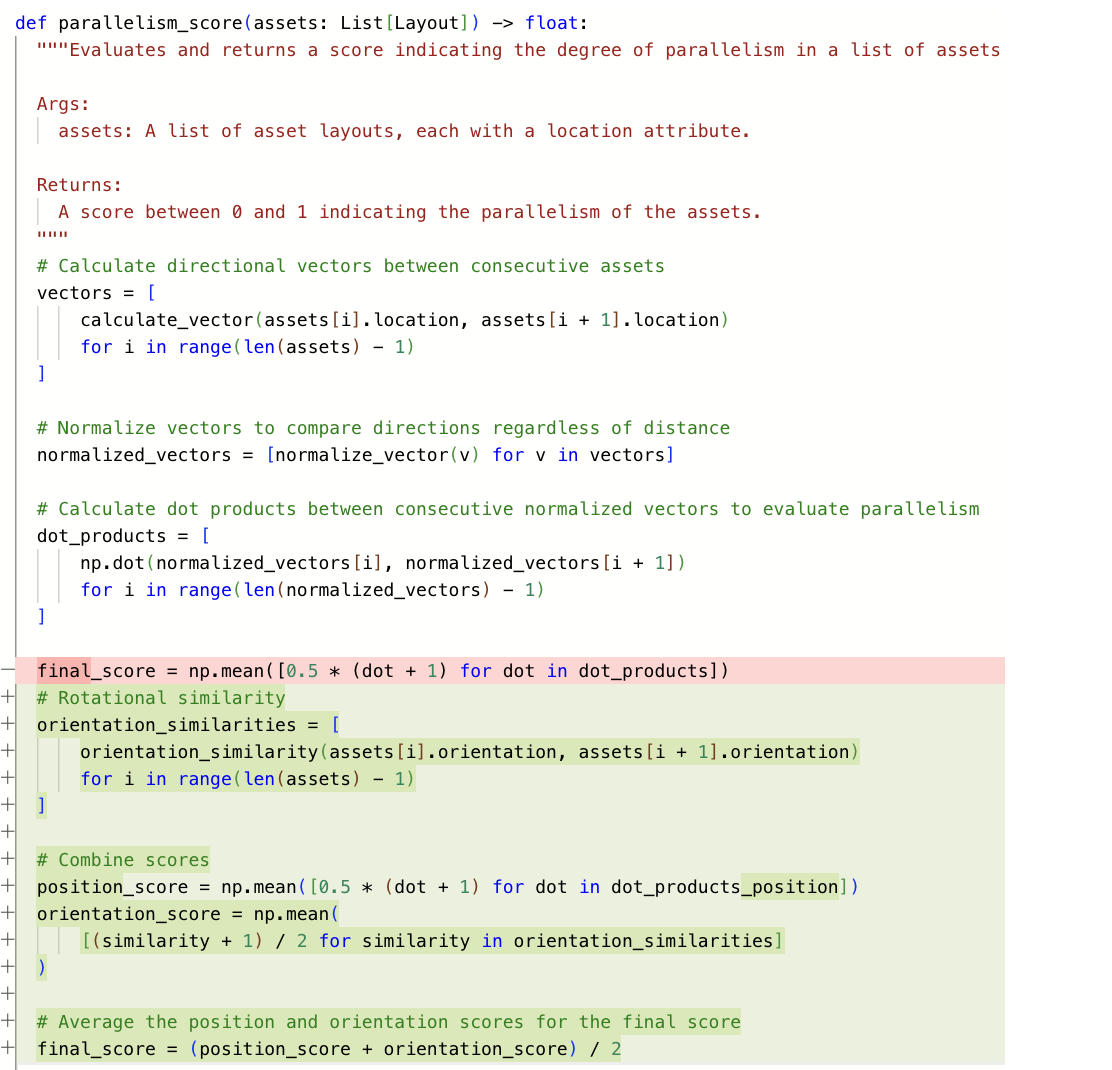}
    \caption{Example of function $\texttt{parallelism\_score}$ update in \textbf{outer-loop library learning} phase. The update adds constraint score forcing the orientation of the assets to be similar.}
    \label{fig:update}
\end{figure}

\begin{algorithm}[h!]
\small
    \caption{Dual-Loop Improvement Workflow}
    \label{alg:inner-outer-loop-learning}
    \SetAlgoLined
    \KwData{$L = \{F_r\}$: Initialize SceneCraft's library, $\mathcal{Q} = \{q\}$: a dataset of queries. $N_{\text{inner}}$ and $N_{\text{outer}}$: number of iterations for scene refinement and library learning.}
    \Repeat(\tcp*[f]{outer-loop}){$N_{\text{outer}}$ times}{
        \For{$q \in \mathcal{Q}$}{ \label{alg:marker}
            Decompose $q$ into sub-scene descriptions and assets: $(q_1, \mathcal{A}_1)\ldots(q_K, \mathcal{A}_K)\gets \texttt{LLM-decomposer}(q)$
            \For{$k = 1$ to $K$}{
                Plan a scene graph $\mathcal{G}(s)$ for sub-scene $q_k$: $\mathcal{G}(s) \gets \texttt{LLM-Planner}(q_k, \mathcal{A}_k)$ 
                
                Write code and arguments for scene $s_k$: $\texttt{code}, \texttt{arg} \gets \texttt{LLM-Coder}(\mathcal{G}(s), q_k)$

                \For(\tcp*[f]{inner-loop}){$t = 1$ to $N_{\text{inner}}$}{
                    Constraint-based Search:\ $\mathcal{L}^t \gets \underset{\mathcal{L}}{\mathrm{argmax}} \sum_{r \in \mathcal{R}} F^t_r\Big(\big\{\mathcal{L}(a_i) \mid a_i \in \mathcal{E}^t(r)\big\}, \texttt{arg}^t_r\Big)$

                    Render: $\texttt{img} \gets \texttt{Blender}(\mathcal{L}^{t}, \mathcal{A}, \texttt{code}^{t})$
                    
                    Critic \& Revise: $\mathcal{E}^{t+1}, \mathcal{F}^{t+1}, \texttt{arg}^{t+1} \gets  \texttt{LLM+V-Reviewer}(\texttt{img}, q_k)$   
                }
            }
            Store function updates $F^T_r(q)$ for each relation $r$ per question $q$ into $\hat{F_r}$ set. 
        }
        \For{$r \in \mathcal{R}$}{
            Update library $L$ with updated functions:
            $F_r \gets \texttt{Library-Learner}\Big(\big\{\hat{F_r}(q) \mid q \in \mathcal{Q}\big\}\Big)$
        }
    }
\end{algorithm}

\subsection{Library Learning} \label{sec:library_learning}
In the preceding sections, we described the methodology behind SceneCraft's generation of scenes, which involves the formulation of relations and constraint scoring functions, followed by their iterative optimization through a feedback loop. In this section, we go over the process by which we learn a comprehensive spatial skill library of constraint functions, designed for re-application in the scene generation process for new input queries.

The core of SceneCraft's library learning originates from the aforementioned self-refinement procedure. When a specific constraint function is not sufficient to cover all cases of a relation, the \texttt{LLM+V-Reviewer} is able to identify the pitfall of function implementation, and make corresponding modification. As an example shown in Figure~\ref{fig:update}, the previous implementation of \textit{parallelism} relation only consider the assets' location. Through feedback-loop optimization for scene improvement, GPT-V identifies that it is necessary to consider the similarity over orientation. 
Therefore, the main goal of library learning procedure is to review  these gradual changes of $F_r$, detect common patterns in the addition or modification, and  merge these changes into the library. 




Specifically, we denote $\hat{F_r}(q) = F_r^{T}(q)$ as the updated functions for relation $r$ learned after $t=T$ step inner-loop self-improvement over query $q$. SceneCraft reviews all these updates, try to find one that represents the consensus of all, and merge it into the global skill library. This procedure shares similar intuition as universal self-consistency~\citep{DBLP:journals/corr/abs-2311-17311}.
We thus learn the new function as:
\begin{align}
F_r \gets \texttt{Library-Learner}\Big(\big\{\hat{F_r}(q) \mid q \in \mathcal{Q} \big\}\Big)
\end{align}
This process is conducted over a batch of queries $\mathcal{Q} = \{q_i\}$ to ensure the universality of the learned skills. Note that: 1) this procedure could be regarded as a meta-learning update of the function initialization to facilitate the feedback-loop optimization. 2) this procedure does not require any ground-truth scenes, explicit reward function, or any human intervention. All the internal learning signal is just the \texttt{LLM+V-reviewer} during the feedback-loop to maximize the alignment to textual query. 3) For both optimization stages, updates are made to non-parametric knowledge represented as Python code, avoiding the computational cost and inaccessibility issues associated with back-propagation in large language models. 

SceneCraft's library learning process is also highly sample-efficient. By manually creating 20 examples with ground-truth constraints and running dual-stage optimization on them, SceneCraft develops a robust skill library. This approach contrasts with traditional model fine-tuning, offering efficiency and adaptability in learning for complex tasks like 3D scene generation. The pseudo-code of the whole dual-loop learning is illustrated in Alg~\ref{alg:inner-outer-loop-learning}.

\section{Experiments}
We evaluate our proposed SceneCraft first on our curated synthetic queries where the ground-truth constraints are available. We then show how the generated 3D scenes can help video generation on the Sintel movie as a case study.


\subsection{Evaluate Scene Synthesis with Given Constraints}

SceneCraft is an agent for open-domain scene synthesis. Most of the existing 3D scene datasets with ground-truth focus on a specific domain such as in-door scene~\citep{DBLP:conf/mm/SongCXKTYY23, DBLP:conf/cvpr/WeiDPSP0G23} or road traffic~\citep{DBLP:journals/corr/abs-2303-08473}.
To systematically study and evaluate our agent in this task, we manually create 40 synthetic queries with ground-truth constraints. The way we generate these queries is by first sampling a subset of relation constraints from the full list (shown in Appendix~\ref{sec:relation}). Based on this, the human annotators evaluate whether the scene satisfies this relational constraint. Assets are retrieved from Turbosquid\footnote{\url{https://www.turbosquid.com/}}.

\paragraph{Evaluation Metric} To verify whether a generated scene fulfills the textual requirement, we ask human annotators to also write a scoring function to estimate how much the constraint is satisfied. Such function is different from the one SceneCraft learns in its skill library, because the scoring function only needs to work for this specific scene query. We show examples of the queries as well as the implemented scoring function in Sec~\ref{sec:constrain} in Appendix. The output of these scoring functions is less than or equal to 1, and only reaches equality when all constraints are strictly satisfied.

For this synthetic dataset, as we don't have the ground-truth scene layout, we adopt two metrics for evaluating scene synthesis model's performance. The first is the standard text-to-image CLIP similarity score~\citep{DBLP:conf/icml/RadfordKHRGASAM21}, which measures how well the generated scene satisfies the textual description; we also use the functions human annotators wrote as a more fine-grained evaluation on how our generated scene satisfies all the semantic requirements hidden in the query. We use 20 of the 40 queries for building the spatial skill library through dual-loop optimization, during which the model only sees the query instead of ground-truth constraint score. Afterwards, we evaluate the model performance on the remaining 20 queries.


\paragraph{Baselines}
Most of the existing 3D scene synthesis works only focus on a specific domain, e.g., indoor scenes. The only prior system that serves similar purpose to SceneCraft might be BlenderGPT\footnote{\url{https://github.com/gd3kr/BlenderGPT}}, an LLM assistant that also takes text query as input and generates Blender code. The main difference of BlenderGPT against SceneCraft is that BlenderGPT is limited to only the basic Blender instructions such as moving an asset or changing texture. To allow it to solve the text-to-scene synthesis task, we modify their code to: 1) enable BlenderGPT to use GPT-V to receive the screenshot of Blender as visual feedback; 2) asking itself to give the per-step instruction for generating the complex scene. 
We also report results of our own system's ablation. There are three major design choices of SceneCraft: 1) Abstraction of scene as relational graph; 2) inner-loop optimization of the scene with visual feedback; 3) outer-loop learning of skill library. 
These three components have some dependencies: constraint function grounded by relational graph is the main interface to be updated by inner-loop (BlenderGPT can be regarded as a baseline only with inner-loop update but without graph grounding); while the inner-loop updates of function is the root for library learning. Therefore, we do ablation study by removing one component after the other.




\paragraph{Experimental Results}
Results are shown in Table~\ref{tab:synthetic}. We see that our method consistently improves over all baselines in terms of both CLIP similarity as well as the constraint score. Notably, on the constraint score, the BlenderGPT baseline only achieves 5.6 score. We show a few head-to-head comparisons in Figure~\ref{fig:teasing}. As example, in the first query that asks three boxes stack one on top of each other, BlenderGPT simply lists the three boxes in a line and does not follow the instruction of stacking; on the second query that asks three trees in a row, BlenderGPT does organize the trees in a line, but perpendicular with the road edge. These examples show that BlenderGPT without the relational constraint is not able to conduct complex spatial planning.

In the meantime, the ablation studies by removing each component also shows that all components are very crucial for Scenecraft. Among these components, inner-loop optimization provides the most important leaps; removing it leads to 38.4 drop on constraint score, and it's also the root for library learning that keep the system self-improving without human annotation.


\begin{table}[t]
\centering
\label{tab:model_comparison}
\vspace{-0.05in}
\resizebox{!}{17mm}{
\begin{tabular}{l|cc}
\toprule
\textbf{Metric} & \textbf{CLIP SIM} & \textbf{  Constraint Score} \\
\midrule
BlenderGPT & 24.7 & 5.6 \\
\textbf{SceneCraft} & \textbf{69.8} & \textbf{88.9}\\ \midrule
\multicolumn{3}{c}{(----Ablation by removing one component after the other----)} \\
~-- Learned Library & 48.3 & 64.5 \\
~~~-- Inner-Loop & 32.8 & 26.1 \\
~~~~~-- Relation Graph & 19.4 & 3.2\\
\bottomrule
\end{tabular}
}
\vspace{-0.15in}
\caption{Comparison of SceneCraft against BlenderGPT and ablation baselines on synthetic queries with annotated constraints. 
}
\label{tab:synthetic}
\vspace{-0.1in}
\end{table}

We also conduct a qualitative  evaluation of SceneCraft's output versus BlenderGPT baseline. We randomly select 10 pairs generated by SceneCraft and BlenderGPT, and ask humans to judge which one is better, in terms of three major dimensions: 1) text fidelity: how much the generated scene aligns with the textual query; 2) composition \& constraint agreement: we tell the raters the ground-truth relations, and ask whether the generated scene follows all these constraints; 3) Aesthetics: we ask which output has better overall visual quality. The order of our output against baseline is completely random. Detailed question and interface is shown in Figure~\ref{fig:question} in Appendix. Altogether we collect 22 responses. 
Results in Table~\ref{tab:human} show that our method outperforms BlenderGPT in all the three dimensions significantly. Specifically, consistent with our results on constraint score, SceneCraft gains more improvement over the constraint agreement, making the scene logically correct.

\begin{table}[t!]
\centering
\label{tab:human_eval}
\resizebox{!}{8mm}{
\begin{tabular}{l|ccc}
\toprule
\textbf{Win Rate} & \textbf{Text Fidelity} & \textbf{Composition} & \textbf{Aesthetics} \\
\midrule
SceneCraft & 76.8\% & 83.6\% & 74.5\%\\
BlenderGPT & 12.7\% & 11.4\%  & 14.5\% \\
\bottomrule
\end{tabular}
}
\vspace{-0.1in}
\caption{Qualitative Human comparison of SceneCraft against BlenderGPT baseline.}
\label{tab:human}
\vspace{-0.2in}
\end{table}

\subsection{Scene-Guided Video Generation over Sintel Movie}
In addition to synthetic queries, we also show that SceneCraft's layout planning capability generalizes to real scenes
, and has potential to control and benefit video generation. As open-domain videos do not always have ground-truth scenes, we take the Sintel Movie, which is an animated fantasy short film produced with Blender, where scripts and Blender scenes are open sourced\footnote{\url{https://studio.blender.org/films/sintel/}}. We download all these scenes, using the first half as the training set and the remaining  half for testing. For this task, we assume that the model is given the ground-truth assets for the scene, and only focuses on layout planning to satisfy the textual description. After we recover the scene, we study how it can benefit a video generation model to get higher-quality predictions. We thus fine-tune the VideoPoet model~\citep{DBLP:journals/corr/abs-2312-14125}, an autoregressive Transformer-based video generation framework, on the training set with one ground-truth scene image frame as a conditional input. The image will be converted into image tokens, and add as prefix after text prompt. We then take the fine-tuned VideoPoet model, taking our model and BlenderGPT's predicted scene, to generate a 2 second video.

\begin{figure}[t!]
    \centering    
    \vspace{-0.1in}
    \includegraphics[width=0.5\textwidth]{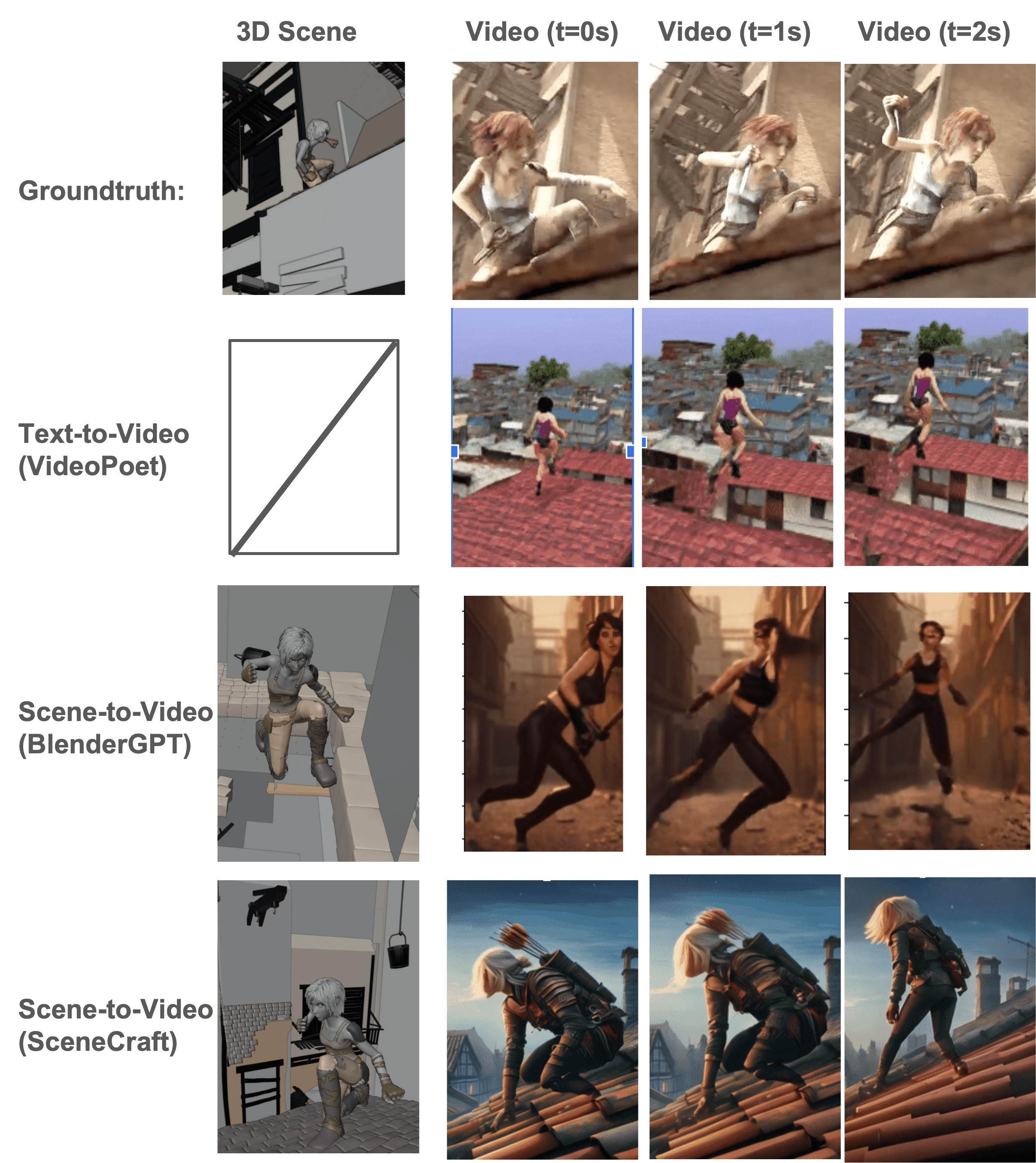}
    \vspace{-0.3in}
    \caption{Predicted 3D Scenes as well as the generated videos by SceneCraft against other baselines. }
    \label{fig:sintel_example}
     \vspace{-0.1in}
\end{figure}

\begin{table*}[t!]
\centering
\vspace{-0.05in}
\resizebox{!}{20mm}{
\begin{tabular}{l|cc|cc}
\toprule
\multirow{2}{*}{Quantitative Metric} & \multicolumn{2}{c|}{\textbf{Scene Comparison} }& \multicolumn{2}{c}{\textbf{Video Comparison}} \\
& \textbf{Layout Matrix SIM} & 
\textbf{Scene CLIP SIM} & \textbf{CLIP-based RM} & \textbf{FVD $\downarrow$} \\
\midrule
Text-to-Video (w.o. / finetune) & / & / & 56.8& 846 \\
Text-to-Video (w / finetune) & / & / & 64.2 & 531 \\ \midrule
\multicolumn{5}{c}{\textbf{Text-to-Scene-to-Video}, finetune a videogen model on groundtruth scene, infer with scenes generated by:} \\ \midrule
BlenderGPT & 27.5 &  41.8 & 69.1 & 574\\
SceneCraft (Dual-Loop) & 69.3 & 82.7 & 46.2 & 317 \\
\bottomrule
\end{tabular}
}
\vspace{-0.05in}
\caption{Comparison of SceneCraft with other ablated baselines on a Sintel movie. In this setting, we assume to be given fixed assets for each scene, try to recover the scene, and guide a video generative model which is fine-tuned on first half of the video. We compare with naiive text-to-video baselines without scene guidance.}
\vspace{-0.2in}
\label{tab:sintel}
\end{table*}

We compare the output in terms of both the scene itself as well as how much it benefits the overall video generation. For the scene, we use two metrics: the layout matrix's similarity (first calculate mutual similarity between assets, then calculate cosine similarity), and the rendered image's CLIP score. For the video, as we use both the standard Frechet Video Distance (FVD) distribution score~\citep{DBLP:conf/iclr/UnterthinerSKMM19}, as well as CLIP-based Relative Matching (RM) score~\citep{DBLP:journals/corr/abs-2104-14806}.
The results shown in Table~\ref{tab:sintel} illustrate that our method consistently improves the BlenderGPT 
output in terms of scene planning. In addition, the generated scene helps the video generation and outperform the vanilla text-to-video baseline. From the examples in Figure~\ref{fig:sintel_example}, we can see that the 3D scene grounding help the generated video follow more similar structure as ground-truth ones. This shows the potential of SceneCraft in controlling video generation in wider domain.


\section{Related Works}
\paragraph{Text to 3D-Scene Synthesis}
One of the earliest forays into text-driven 3D scene synthesis is WordsEye~\citep{DBLP:conf/siggraph/CoyneS01}. This system, and its follow-up works~\citep{DBLP:conf/mm/SeverskyY06, DBLP:conf/emnlp/ChangSM14, DBLP:journals/tog/MaPFLPHYTGZ18}, can generate 3D scenes from natural language. However, these systems often require manual mapping between language and object placement, leading to somewhat unnatural commands for scene description. \citet{DBLP:conf/iccv/ZitnickPV13} learns to map visual features to semantic phrases extracted from sentences, focusing on binary spatial or semantic relationships. 
\citet{DBLP:conf/emnlp/ChangSM14} build upon and improve these early systems. The key advancement is the use of spatial knowledge, derived from 3D scene data, to more accurately constrain scene generations. This approach allows for a more realistic interpretation of unstated facts or common sense in scene synthesis. In their subsequent work~\citep{DBLP:conf/acl/ChangMSPM15}, they focused on lexical grounding of textual terms to 3D model references, combining rule-based models with user annotations to select appropriate objects. Their latest paper~\citep{DBLP:journals/corr/ChangESM17} further refines this approach, introducing interactive text-based scene editing operations and an improved user interface. All these systems are most purely symbolic rule-based and require significant human efforts to maintain, and are, therefore, hard to generalize to new domains and types of constraints.

There also exist a line of neural-based 3D scene generation that learns from data. Most works in this direction focus on a specific domain, such as in-door scenes~\citep{DBLP:journals/corr/abs-2304-03188}. For instance, RoomDreamer~\citep{DBLP:conf/mm/SongCXKTYY23} trains a diffusion model to simultaneously generate layout, geometry and texture for in-door scenes; LEGO-Net~\citep{DBLP:conf/cvpr/WeiDPSP0G23} focus on the layout planning, and trains a Transformer model to iteratively cleanup the messy room. Despite the impressive performance of these work, they are restricted by the available 3D scene data. For most open-domain image and videos, it is very hard to collect ground-truth 3D scenes, which is why most works in this domain focus on in-door scenes. On the contrary, this paper focus on exploring whether we can take advantage of the existing knowledge and reasoning capabilities of Large Language Models to directly do layout planning without tuning its parameters, and we try to learn general spatial planning skills that can be generalized from very small number of synthetic queries.

\paragraph{Multimodal LLM Agents}
Leverageing visual perception abilities of recent models like GPT-V, multimodal LLM Agents~\citep{DBLP:journals/corr/abs-2308-03688} are capable of interacting with external visual environments, such as web browsing~\citep{DBLP:journals/corr/abs-2306-06070, zhou2023webarena, DBLP:journals/corr/abs-2306-08129, zheng2023seeact}, gaming~\citep{DBLP:journals/corr/abs-2305-16291}, robotics~\citep{DBLP:journals/corr/abs-2307-15818} and design~\citep{lv2023gpt4motion, yang2024mastering}. The most related concurrent works is 3D-GPT~\citep{DBLP:journals/corr/abs-2310-12945}, which interacts with Infinigen~\citep{DBLP:conf/cvpr/RaistrickLMMWZK23}, a high-level wrapper on top of Blender, to create high-quality environmental scenes. The main difference of our work against 3D-GPT\footnote{The code of this work hasn't released, and we plan to compare after they open-source the code.} is: 1) Environment-wise, we directly interact with Blender and a large-scale asset pool, which provide richer assets to construct the scene, while Infinigen for now only supports limited number of assets and environment arguments; 2) methodology-wise, SceneCraft features a dual-loop self-improvement pipeline, which enables us to learn new design skills to handle unseen tasks, which differentiate us to many existing llm-agent works that heavily rely on manual prompt design.

\section{Conclusion}
In this paper, we present SceneCraft, an LLM-powered autonomous agent for transforming input text query to a 3D Scene by generating a Blender-executable Python script. Scenecraft builds on top of multimodal LLMs for both planning and library learning in a dual-loop self-improving framework. In the inner-loop, SceneCraft generates Blender-executable Python scripts to render an image of the scene, and use a Self-critiquing loop to iteratively refine its output and learn from its performance. The outer-loop dynamically expands a 'spatial skill' library, 
facilitating continuous self-improvement without the need for expensive LLM parameter tuning. In the future, we'd like to explore: 1) using our framework for reconstructing the 3D scene corresponding to a given open-domain image or video; 2) utilizing the generated dataset to fine-tune a video generation conditioned on a 3D scene as control signal.

\section*{Impact Statements}
This paper presents work whose goal is to advance the text-to-3d-scene synthesis. The work can have potential to benefit the gaming, cinematic and design industry, which are mostly positive impact. We only learn a non-parametric skill library from synthetic queries, and not using any private information. There are many potential societal consequences of our work, none which we feel must be specifically highlighted here.

\bibliography{main}
\bibliographystyle{icml2023}

\clearpage
\appendix
\onecolumn
\begin{center}
	{\Large \textbf{Supplementary Material for \textsc{SceneCraft}}}
\end{center}
\section{Examples of SceneCraft's Generated Scripts and Rendered Scenes}\label{sec:example}
\begin{figure*}
    \centering
    \includegraphics[width=0.8\textwidth]{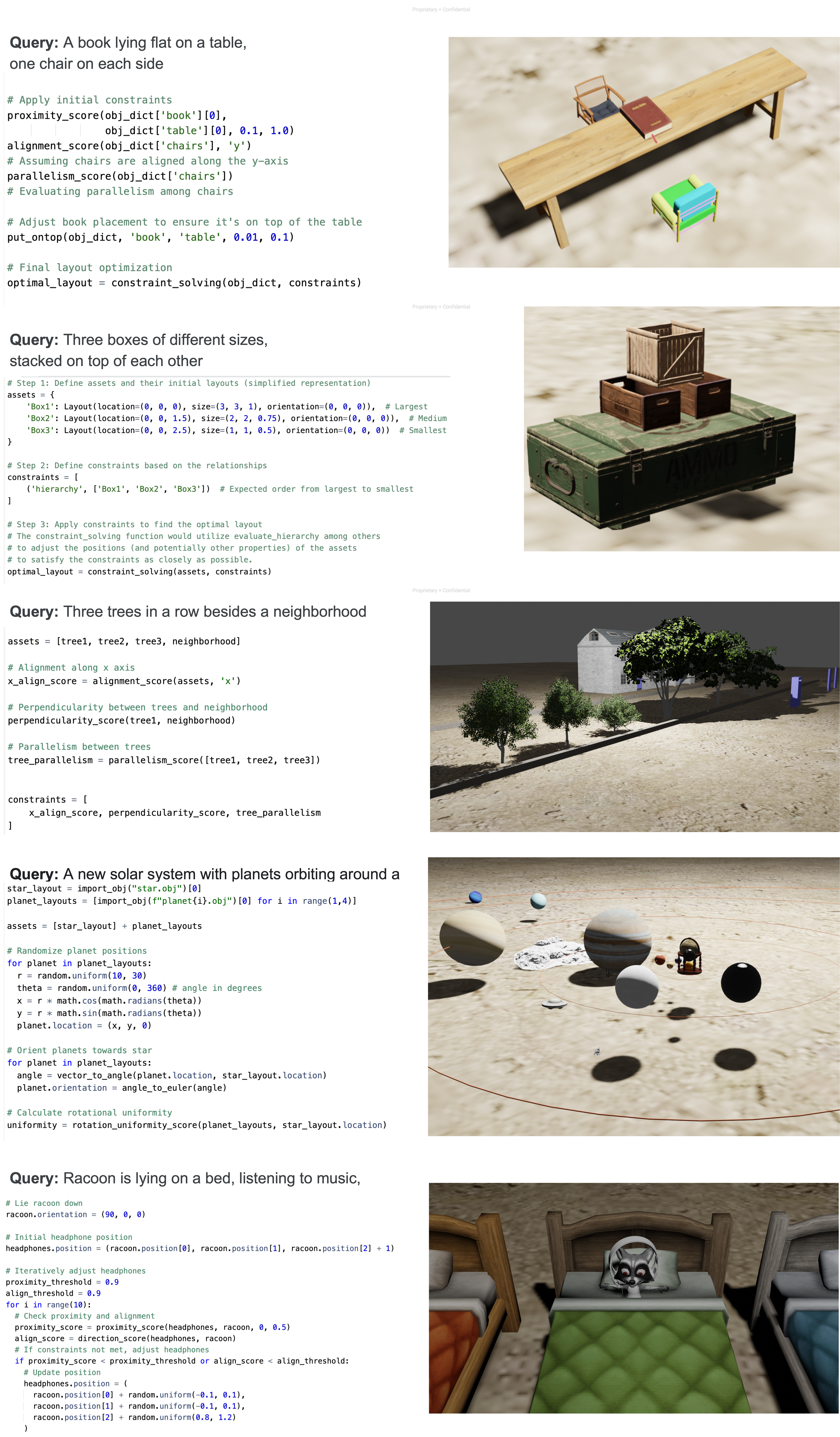}
    \caption{Examples of generated code and scenes}
    \label{fig:example_1}
\end{figure*}

\begin{figure*}
    \centering
    \includegraphics[width=1.0\textwidth]{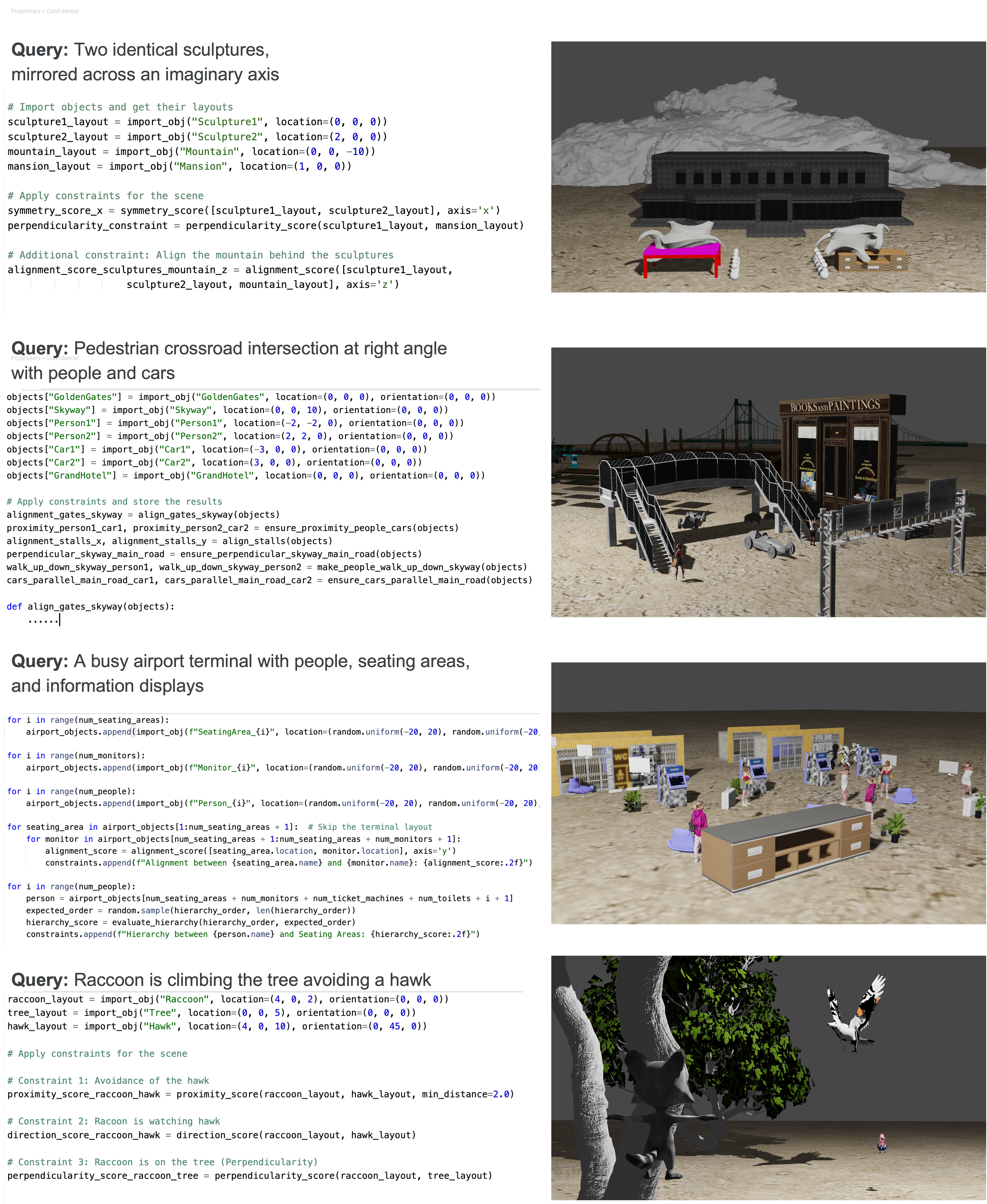}
    \caption{Examples of generated code and scenes}
    \label{fig:example_2}
\end{figure*}

Example of SceneCraft's generated scripts and rendered scene on the synthetic datasets are in Figure~\ref{fig:example_1} and Figure~\ref{fig:example_2}.

\section{List of relationships} \label{sec:relation}
SceneCraft encapsulates several types of relationships and constraints, including:
\begin{compactitem}
    \item \textbf{Proximity}: A constraint enforcing the closeness of two objects, e.g., a chair near a table.
    \item \textbf{Direction}: The angle of one object is targeting at the other.
    \item \textbf{Alignment}: Ensuring objects align along a common axis, e.g., paintings aligned vertically on a wall.
    \item \textbf{Symmetry}: Mirroring objects along an axis, e.g., symmetrical placement of lamps on either side of a bed.
    \item \textbf{Overlap}: One object partially covering another, creating depth, e.g., a rug under a coffee table.
    \item \textbf{Parallelism}: Objects parallel to each other, suggesting direction, e.g., parallel rows of seats in a theater.
    \item \textbf{Perpendicularity}: Objects intersecting at a right angle, e.g., a bookshelf perpendicular to a desk.
    \item \textbf{Hierarchy}: Indicating a list of objects follow a certain order of size / volumns.
    \item \textbf{Rotation}: a list of objects rotate a cirtain point, e.g., rotating chairs around a meeting table.
    \item \textbf{Repetition}: Repeating patterns for rhythm or emphasis, e.g., a sequence of street lights.
    \item \textbf{Scaling}: Adjusting object sizes for depth or focus, e.g., smaller background trees to create depth perception.
\end{compactitem}

These relationships are vital for creating scenes that are not only visually appealing but also contextually coherent. Traditionally the functions $F(\cdot)$ for each constraint shall be written by human experts, and SceneCraft's major contribution is to autonomously learn and evolve the library of constraint satisfaction functions $\mathcal{F} = \{{F_c(\cdot)}\}_{c\in \mathcal{C}}$, using a Large Language Model (LLM) Agent.

\section{Spatial Skill Library}
Below listed all the functions our framework generate. There exist some basic fundamental editing functions like import object, add camera, scaling, repetition; some functions to get information from the scene, such as calculate shortest distance between objects, calculate volumn, etc; as well as functions that calculate constraint satisfying score for each relationship. All these functions are autonomously written and modified by LLM Agent itself, without ground-truth label or explicit human intervention:

\begin{figure*}[!ht] 
\begin{lstlisting}[language=Python,label={code:task},floatplacement=H]
@dataclass
class Layout:
    location: Tuple[float, float, float]
    min: Tuple[float, float, float]
    max: Tuple[float, float, float]
    orientation: Tuple[float, float, float] # Euler angles (pitch, yaw, roll)

def scale_group(objects: List[bpy.types.Object], scale_factor: float) -> None:
    """
    Scale a group of objects by a given factor.

    Args:
        objects (List[bpy.types.Object]): List of Blender objects to scale.
        scale_factor (float): The scale factor to apply.

    Example:
        scale_group([object1, object2], 1.5)
    """
    for obj in objects:
        obj.scale = (obj.scale.x * scale_factor, 
                     obj.scale.y * scale_factor, 
                     obj.scale.z * scale_factor)
        obj.matrix_world = obj.matrix_world * scale_factor

def find_highest_vertex_point(objs: List[bpy.types.Object]) -> Dict[str, float]:
    """
    Find the highest vertex point among a list of objects.

    Args:
        objs (List[bpy.types.Object]): List of Blender objects to evaluate.

    Returns:
        Dict[str, float]: The lowest x, y, and z coordinates.

    Example:
        lowest_point = find_lowest_vertex_point([object1, object2])
    """
    bpy.context.view_layer.update()
    highest_points = {'x': -float('inf'), 'y': -float('inf'), 'z': -float('inf')}

    for obj in objs:
        # Apply the object's current transformation to its vertices
        obj_matrix_world = obj.matrix_world

        if obj.type == 'MESH':
            # Update mesh to the latest data
            obj.data.update()
            for vertex in obj.data.vertices:
                world_vertex = obj_matrix_world @ vertex.co
                highest_points['x'] = max(highest_points['x'], world_vertex.x)
                highest_points['y'] = max(highest_points['y'], world_vertex.y)
                highest_points['z'] = max(highest_points['z'], world_vertex.z)
        
    return highest_points

def find_lowest_vertex_point(objs: List[bpy.types.Object]) -> Dict[str, float]:
    """
    Find the lowest vertex point among a list of objects.

    Args:
        objs (List[bpy.types.Object]): List of Blender objects to evaluate.

    Returns:
        Dict[str, float]: The lowest x, y, and z coordinates.

    Example:
        lowest_point = find_lowest_vertex_point([object1, object2])
    """
    bpy.context.view_layer.update()
    lowest_points = {'x': float('inf'), 'y': float('inf'), 'z': float('inf')}

    for obj in objs:
        # Apply the object's current transformation to its vertices
        obj_matrix_world = obj.matrix_world

        if obj.type == 'MESH':
            # Update mesh to the latest data
            obj.data.update()
            for vertex in obj.data.vertices:
                world_vertex = obj_matrix_world @ vertex.co
                lowest_points['x'] = min(lowest_points['x'], world_vertex.x)
                lowest_points['y'] = min(lowest_points['y'], world_vertex.y)
                lowest_points['z'] = min(lowest_points['z'], world_vertex.z)
        
    return lowest_points
\end{lstlisting}
\end{figure*}

\begin{figure*}[!ht] 
\begin{lstlisting}[language=Python,label={code:task},floatplacement=H]
def rotate_objects_z_axis(objects: List[bpy.types.Object], angle_degrees: float) -> None:
    """
    Rotate a group of objects around the Z-axis by a given angle.

    Args:
        objects (List[bpy.types.Object]): List of objects to rotate.
        angle_degrees (float): The angle in degrees to rotate.

    Example:
        rotate_objects_z_axis([object1, object2], 45)
    """
    bpy.context.view_layer.update()
    angle_radians = math.radians(angle_degrees)  # Convert angle to radians
    rotation_matrix = mathutils.Matrix.Rotation(angle_radians, 4, 'Y')
    lowest_point = find_lowest_vertex_point(objects)
    highest_points = find_highest_vertex_point(objects)
    center_point = {'x': (lowest_point['x'] + highest_points['x']) / 2, 
                    'y': (lowest_point['y'] + highest_points['y']) / 2,
                    'z': 0}
    for obj in objects:
        if obj.type == 'MESH':
            obj.data.update()
            obj.matrix_world = obj.matrix_world @ rotation_matrix
            
    lowest_point = find_lowest_vertex_point(objects)
    highest_points = find_highest_vertex_point(objects)
    center_point['x'] -= (lowest_point['x'] + highest_points['x']) / 2 
    center_point['y'] -= (lowest_point['y'] + highest_points['y']) / 2 
    shift(objects, center_point)
    
def shift(objects: List[bpy.types.Object], shift_loc: Dict[str, float]) -> None:
    """
    Shift a group of objects with shift_loc.

    Args:
        objects (List[bpy.types.Object]): List of objects to rotate.
        shift_loc (float): The shift vector.

    Example:
        rotate_objects_z_axis([object1, object2], (5,3,1))
    """
    for obj in objects:
        # Shift object so the lowest point is at (0,0,0)
        obj.location.x += shift_loc['x']
        obj.location.y += shift_loc['y']
        obj.location.z += shift_loc['z']
    bpy.context.view_layer.update()

def calculate_shortest_distance(vertices1: Set[Tuple[float, float, float]], vertices2: Set[Tuple[float, float, float]]) -> float:
    """
    Calculate the shortest distance between two sets of vertices.

    Args:
        vertices1 (Set[Tuple[float, float, float]]): First set of vertices.
        vertices2 (Set[Tuple[float, float, float]]): Second set of vertices.

    Returns:
        float: Shortest distance over the Z-axis.
    """
    min_distance = float('inf')
    for v1_tuple in vertices1:
        v1 = Vector(v1_tuple)
        for v2_tuple in vertices2:
            v2 = Vector(v2_tuple)
            distance = (v1 - v2).length
            min_distance = min(min_distance, distance)
    return min_distance



\end{lstlisting}
\end{figure*}

\begin{figure*}[!ht] 
\begin{lstlisting}[language=Python,label={code:task},floatplacement=H]
def rotate_objects_z_axis(objects: List[bpy.types.Object], angle_degrees: float) -> None:
    """
    Rotate a group of objects around the Z-axis by a given angle.

    Args:
        objects (List[bpy.types.Object]): List of objects to rotate.
        angle_degrees (float): The angle in degrees to rotate.

    Example:
        rotate_objects_z_axis([object1, object2], 45)
    """
    bpy.context.view_layer.update()
    angle_radians = math.radians(angle_degrees)  # Convert angle to radians
    rotation_matrix = mathutils.Matrix.Rotation(angle_radians, 4, 'Y')
    lowest_point = find_lowest_vertex_point(objects)
    highest_points = find_highest_vertex_point(objects)
    center_point = {'x': (lowest_point['x'] + highest_points['x']) / 2, 
                    'y': (lowest_point['y'] + highest_points['y']) / 2,
                    'z': 0}
    for obj in objects:
        if obj.type == 'MESH':
            obj.data.update()
            obj.matrix_world = obj.matrix_world @ rotation_matrix
            
    lowest_point = find_lowest_vertex_point(objects)
    highest_points = find_highest_vertex_point(objects)
    center_point['x'] -= (lowest_point['x'] + highest_points['x']) / 2 
    center_point['y'] -= (lowest_point['y'] + highest_points['y']) / 2 
    shift(objects, center_point)
    
def shift(objects: List[bpy.types.Object], shift_loc: Dict[str, float]) -> None:
    """
    Shift a group of objects with shift_loc.

    Args:
        objects (List[bpy.types.Object]): List of objects to rotate.
        shift_loc (float): The shift vector.

    Example:
        rotate_objects_z_axis([object1, object2], (5,3,1))
    """
    for obj in objects:
        # Shift object so the lowest point is at (0,0,0)
        obj.location.x += shift_loc['x']
        obj.location.y += shift_loc['y']
        obj.location.z += shift_loc['z']
    bpy.context.view_layer.update()

def calculate_shortest_distance(vertices1: Set[Tuple[float, float, float]], vertices2: Set[Tuple[float, float, float]]) -> float:
    """
    Calculate the shortest distance between two sets of vertices.

    Args:
        vertices1 (Set[Tuple[float, float, float]]): First set of vertices.
        vertices2 (Set[Tuple[float, float, float]]): Second set of vertices.

    Returns:
        float: Shortest distance over the Z-axis.
    """
    min_distance = float('inf')
    for v1_tuple in vertices1:
        v1 = Vector(v1_tuple)
        for v2_tuple in vertices2:
            v2 = Vector(v2_tuple)
            distance = (v1 - v2).length
            min_distance = min(min_distance, distance)
    return min_distance



\end{lstlisting}
\end{figure*}

\begin{figure*}[!ht] 
\begin{lstlisting}[language=Python,label={code:task},floatplacement=H]

def check_vertex_overlap(vertices1: Set[Vector], vertices2: Set[Vector], threshold: float = 0.01) -> float:
    """
    Check if there is any overlap between two sets of vertices within a threshold.

    Args:
        vertices1 (Set[Vector]): First set of vertices.
        vertices2 (Set[Vector]): Second set of vertices.
        threshold (float): Distance threshold to consider as an overlap.

    Returns:
        bool: True if there is an overlap, False otherwise.
    """
    for v1_tuple in vertices1:
        v1 = Vector(v1_tuple)
        for v2_tuple in vertices2:
            v2 = Vector(v2_tuple)
            if (v1 - v2).length <= threshold:
                return 1.0
    return 0.0
    
def evaluate_constraints(assets, constraints):
    """Evaluate all constraints and return the overall score."""
    total_score = 0
    for constraint_func, involved_assets in constraints:
        # Assuming each constraint function takes involved assets and returns a score
        scores = constraint_func([assets[name] for name in involved_assets])
        total_score += sum(scores)  # Summing scores assuming each constraint can contribute multiple scores
    return total_score

def adjust_positions(assets, adjustment_step=0.1):
    """Randomly adjust the positions of assets."""
    for asset in assets.values():
        # Randomly adjust position within a small range to explore the space
        asset.location = (
            asset.location[0] + random.uniform(-adjustment_step, adjustment_step),
            asset.location[1] + random.uniform(-adjustment_step, adjustment_step),
            asset.location[2]  # Z position kept constant for simplicity
        )

def constraint_solving(assets, constraints, max_iterations=100):
    """Find an optimal layout of assets to maximize the score defined by constraints."""
    best_score = evaluate_constraints(assets, constraints)
    best_layout = {name: asset.copy() for name, asset in assets.items()}  # Assuming a copy method exists

    for _ in range(max_iterations):
        adjust_positions(assets)
        current_score = evaluate_constraints(assets, constraints)
        
        if current_score > best_score:
            best_score = current_score
            best_layout = {name: asset.copy() for name, asset in assets.items()}
        else:
            # Revert to best layout if no improvement
            assets = {name: layout.copy() for name, layout in best_layout.items()}

    return best_layout, best_score
\end{lstlisting}
\end{figure*}
    
\begin{figure*}[!ht] 
\begin{lstlisting}[language=Python,label={code:task},floatplacement=H]

def normalize_vector(v: np.ndarray) -> np.ndarray:
    """Normalize a vector."""
    norm = np.linalg.norm(v)
    return v / norm if norm > 0 else np.zeros_like(v)

def orientation_similarity(orientation1: Tuple[float, float, float], orientation2: Tuple[float, float, float]) -> float:
    """Calculate the similarity between two orientations, represented as Euler angles."""
    # Convert Euler angles to vectors for simplicity in comparison
    vector1 = np.array(orientation1)
    vector2 = np.array(orientation2)
    # Calculate the cosine similarity between the two orientation vectors
    cos_similarity = np.dot(vector1, vector2) / (np.linalg.norm(vector1) * np.linalg.norm(vector2))
    return cos_similarity

def parallelism_score(assets: List[Layout]) -> float:
    """
    Evaluates and returns a score indicating the degree of parallelism in a list of assets' layouts, considering both position and orientation.
    
    Args:
    assets (List[Layout]): A list of asset layouts.
    
    Returns:
    float: A score between 0 and 1 indicating the parallelism of the assets.
    """
    if len(assets) < 2:
        return 1.0  # Single asset or no asset is arbitrarily considered perfectly parallel
    
    # Positional parallelism
    vectors = [calculate_vector(assets[i].location, assets[i+1].location) for i in range(len(assets)-1)]
    normalized_vectors = [normalize_vector(v) for v in vectors]
    dot_products_position = [np.dot(normalized_vectors[i], normalized_vectors[i+1]) for i in range(len(normalized_vectors)-1)]
    
    # Rotational similarity
    orientation_similarities = [orientation_similarity(assets[i].orientation, assets[i+1].orientation) for i in range(len(assets)-1)]
    
    # Combine scores
    position_score = np.mean([0.5 * (dot + 1) for dot in dot_products_position])
    orientation_score = np.mean([(similarity + 1) / 2 for similarity in orientation_similarities])
    
    # Average the position and orientation scores for the final score
    final_score = (position_score + orientation_score) / 2
    
    return final_score


def calculate_distance(location1: Tuple[float, float, float], location2: Tuple[float, float, float]) -> float:
    """Calculate the Euclidean distance between two points."""
    return np.linalg.norm(np.array(location1) - np.array(location2))

def proximity_score(object1: Layout, object2: Layout, min_distance: float = 1.0, max_distance: float = 5.0) -> float:
    """
    Calculates a proximity score indicating how close two objects are, with 1 being very close and 0 being far apart.
    
    Args:
    object1 (Layout): The first object's layout.
    object2 (Layout): The second object's layout.
    min_distance (float): The distance below which objects are considered to be at optimal closeness. Scores 1.
    max_distance (float): The distance beyond which objects are considered too far apart. Scores 0.
    
    Returns:
    float: A score between 0 and 1 indicating the proximity of the two objects.
    """
    distance = calculate_distance(object1.location, object2.location)
    
    if distance <= min_distance:
        return 1.0
    elif distance >= max_distance:
        return 0.0
    else:
        # Linearly interpolate the score based on the distance
        return 1 - (distance - min_distance) / (max_distance - min_distance)


\end{lstlisting}
\end{figure*}

\begin{figure*}[!ht] 
\begin{lstlisting}[language=Python,label={code:task},floatplacement=H]

def euler_to_forward_vector(orientation: Tuple[float, float, float]) -> np.ndarray:
    """Convert Euler angles to a forward direction vector."""
    pitch, yaw, _ = orientation
    # Assuming the angles are in radians
    x = np.cos(yaw) * np.cos(pitch)
    y = np.sin(yaw) * np.cos(pitch)
    z = np.sin(pitch)
    return np.array([x, y, z])

def calculate_vector(a: Tuple[float, float, float], b: Tuple[float, float, float]) -> np.ndarray:
    """Calculate the directional vector from point a to b."""
    return np.array(b) - np.array(a)

def direction_score(object1: Layout, object2: Layout) -> float:
    """
    Calculates a score indicating how directly object1 is targeting object2.
    
    Args:
    object1 (Layout): The first object's layout, assumed to be the one doing the targeting.
    object2 (Layout): The second object's layout, assumed to be the target.
    
    Returns:
    float: A score between 0 and 1 indicating the directionality of object1 towards object2.
    """
    forward_vector = euler_to_forward_vector(object1.orientation)
    target_vector = calculate_vector(object1.location, object2.location)
    # Normalize vectors to ensure the dot product calculation is based only on direction
    forward_vector_normalized = normalize_vector(forward_vector)
    target_vector_normalized = normalize_vector(target_vector)
    # Calculate the cosine of the angle between the two vectors
    cos_angle = np.dot(forward_vector_normalized, target_vector_normalized)
    # Map the cosine range [-1, 1] to a score range [0, 1]
    score = (cos_angle + 1) / 2
    return score

def alignment_score(assets: List[Layout], axis: str) -> float:
    """
    Calculates an alignment score for a list of assets along a specified axis.
    
    Args:
    assets (List[Layout]): A list of asset layouts to be evaluated for alignment.
    axis (str): The axis along which to evaluate alignment ('x', 'y', or 'z').
    
    Returns:
    float: A score between 0 and 1 indicating the degree of alignment along the specified axis.
    """
    if not assets or axis not in ['x', 'y', 'z']:
        return 0.0  # Return a score of 0 for invalid input
    
    # Axis index mapping to the location tuple
    axis_index = {'x': 0, 'y': 1, 'z': 2}[axis]
    
    # Extract the relevant coordinate for each asset based on the chosen axis
    coordinates = [asset.location[axis_index] for asset in assets]
    # Calculate the variance of these coordinates
    variance = np.var(coordinates)
    # Inverse the variance to calculate the score, assuming a lower variance indicates better alignment
    # Normalize the score to be between 0 and 1, considering a reasonable threshold for "perfect" alignment
    threshold_variance = 1.0  # Define a threshold variance for "perfect" alignment
    score = 1 / (1 + variance / threshold_variance)
    # Clamp the score between 0 and 1
    score = max(0, min(score, 1))
    return score
    
def check_vertex_overlap(vertices1: Set[Vector], vertices2: Set[Vector], threshold: float = 0.01) -> float:
    """
    Check if there is any overlap between two sets of vertices within a threshold.

    Args:
        vertices1 (Set[Vector]): First set of vertices.
        vertices2 (Set[Vector]): Second set of vertices.
        threshold (float): Distance threshold to consider as an overlap.

    Returns:
        bool: True if there is an overlap, False otherwise.
    """
    for v1_tuple in vertices1:
        v1 = Vector(v1_tuple)
        for v2_tuple in vertices2:
            v2 = Vector(v2_tuple)
            if (v1 - v2).length <= threshold:
                return 0.0
    return 1.0
    
\end{lstlisting}
\end{figure*}

\begin{figure*}[!ht] 
\begin{lstlisting}[language=Python,label={code:task},floatplacement=H]


\end{lstlisting}
\end{figure*}

\begin{figure*}[h!]
\begin{lstlisting}[language=Python,label={code:task},floatplacement=H]
def symmetry_score(assets: List[Layout], axis: str) -> float:
    """
    Calculates a symmetry score for a list of assets along a specified axis.
    
    Args:
    assets (List[Layout]): A list of asset layouts to be evaluated for symmetry.
    axis (str): The axis along which to evaluate symmetry ('x', 'y', or 'z').
    
    Returns:
    float: A score between 0 and 1 indicating the degree of symmetry along the specified axis.
    """
    if not assets or axis not in ['x', 'y', 'z']:
        return 0.0  # Return a score of 0 for invalid input
    
    # Axis index mapping to the location tuple
    axis_index = {'x': 0, 'y': 1, 'z': 2}[axis]
    
    # Find the median coordinate along the specified axis to define the symmetry axis
    coordinates = [asset.location[axis_index] for asset in assets]
    symmetry_axis = np.median(coordinates)
    
    # Calculate the deviation from symmetry for each asset
    deviations = []
    for asset in assets:
        # Find the mirrored coordinate across the symmetry axis
        mirrored_coordinate = 2 * symmetry_axis - asset.location[axis_index]
        # Find the closest asset to this mirrored coordinate
        closest_distance = min(abs(mirrored_coordinate - other.location[axis_index]) for other in assets)
        deviations.append(closest_distance)
    
    # Calculate the average deviation from perfect symmetry
    avg_deviation = np.mean(deviations)
    
    # Convert the average deviation to a score, assuming smaller deviations indicate better symmetry
    # The scoring formula can be adjusted based on the specific requirements for symmetry in the application
    max_deviation = 10.0  # Define a maximum deviation for which the score would be 0
    score = max(0, 1 - avg_deviation / max_deviation)
    
    return score

def perpendicularity_score(object1: Layout, object2: Layout) -> float:
    """
    Calculates a score indicating how perpendicular two objects are, based on their forward direction vectors.
    
    Args:
    object1 (Layout): The first object's layout, including its orientation as Euler angles.
    object2 (Layout): The second object's layout, including its orientation as Euler angles.
    
    Returns:
    float: A score between 0 and 1 indicating the degree of perpendicularity.
    """
    vector1 = euler_to_forward_vector(object1.orientation)
    vector2 = euler_to_forward_vector(object2.orientation)
    cos_angle = np.dot(vector1, vector2) / (np.linalg.norm(vector1) * np.linalg.norm(vector2))
    score = 1 - np.abs(cos_angle)
    return score

def calculate_volume(layout: Layout) -> float:
    """Calculate the volume of an object based on its layout dimensions."""
    length = abs(layout.max[0] - layout.min[0])
    width = abs(layout.max[1] - layout.min[1])
    height = abs(layout.max[2] - layout.min[2])
    return length * width * height


\end{lstlisting}
\end{figure*}

\begin{figure*}[h!]
\begin{lstlisting}[language=Python,label={code:task},floatplacement=H]

def evaluate_hierarchy(assets: List[Layout], expected_order: List[str]) -> float:
    """
    Evaluates how well a list of objects follows a specified hierarchical order based on size.
    
    Args:
    assets (List[Layout]): A list of asset layouts to be evaluated.
    expected_order (List[str]): A list of identifiers (names) for the assets, specifying the expected order of sizes.
    
    Returns:
    float: A metric indicating how well the actual sizes of the objects match the expected hierarchical order.
    """
    # Map identifiers to volumes
    id_to_volume = {asset_id: calculate_volume(asset) for asset_id, asset in zip(expected_order, assets)}
    
    # Calculate the actual order based on sizes
    actual_order = sorted(id_to_volume.keys(), key=lambda x: id_to_volume[x], reverse=True)
    
    # Evaluate the match between the expected and actual orders
    correct_positions = sum(1 for actual, expected in zip(actual_order, expected_order) if actual == expected)
    total_positions = len(expected_order)
    
    # Calculate the match percentage as a measure of hierarchy adherence
    match_percentage = correct_positions / total_positions
    
    return match_percentage


def calculate_angle_from_center(center: Tuple[float, float, float], object_location: Tuple[float, float, float]) -> float:
    """Calculate the angle of an object relative to a central point."""
    vector = np.array(object_location) - np.array(center)
    angle = np.arctan2(vector[1], vector[0])
    return angle

def rotation_uniformity_score(objects: List[Layout], center: Tuple[float, float, float]) -> float:
    """
    Calculates how uniformly objects are distributed around a central point in terms of rotation.
    
    Args:
    objects (List[Layout]): A list of object layouts to be evaluated.
    center (Tuple[float, float, float]): The central point around which objects are rotating.
    
    Returns:
    float: A score between 0 and 1 indicating the uniformity of object distribution around the center.
    """
    angles = [calculate_angle_from_center(center, obj.location) for obj in objects]
    angles = np.sort(np.mod(angles, 2*np.pi))  # Normalize angles to [0, 2\pi] and sort
    
    # Calculate differences between consecutive angles, including wrap-around difference
    angle_diffs = np.diff(np.append(angles, angles[0] + 2*np.pi))
    
    # Evaluate uniformity as the variance of these differences
    variance = np.var(angle_diffs)
    uniformity_score = 1 / (1 + variance)  # Inverse variance, higher score for lower variance
    
    return uniformity_score

def put_ontop(obj_dict, moving_set_name, target_set_name, threshold, step):
    """
    Adjust objects in moving_set_name until the shortest distance to target_set_name is below the threshold.

    Args:
        obj_dict (dict): Dictionary of object sets.
        moving_set_name (str): The key for the set of objects to move.
        target_set_name (str): The key for the set of objects to calculate distance to.
        threshold (float): The distance threshold.
        step (float): The step by which to move objects in the Z direction.
    """
    while True:
        vertices_set1 = get_all_vertices(obj_dict[moving_set_name])
        vertices_set2 = get_all_vertices(obj_dict[target_set_name])
        shortest_distance = calculate_shortest_distance(vertices_set1, vertices_set2)
        print(shortest_distance)

        if shortest_distance < threshold:
            break

        for obj in obj_dict[moving_set_name]:
            obj.location.z -= max(step, shortest_distance)

        bpy.context.view_layer.update()
\end{lstlisting}
\end{figure*}

\begin{figure*}[h!]\label{fig:gen_2}
\begin{lstlisting}[language=Python,label={code:task},floatplacement=H]
def repeat_object(original: Layout, direction: Tuple[float, float, float], repetitions: int, distance: float) -> List[Layout]:
    """
    Creates a series of duplicated objects based on the original, repeating them in a specified direction at a set distance.
    
    Args:
    original (Layout): The original object to be repeated.
    direction (Tuple[float, float, float]): The direction vector along which to repeat the object.
    repetitions (int): The number of times the object should be repeated.
    distance (float): The distance between each object.
    
    Returns:
    List[Layout]: A list of Layout objects representing the original and its duplicates.
    """
    repeated_objects = [original]  # Include the original object in the output list
    
    for i in range(1, repetitions):
        # Calculate the new location for each repeated object
        new_location = (
            original.location[0] + direction[0] * distance * i,
            original.location[1] + direction[1] * distance * i,
            original.location[2] + direction[2] * distance * i
        )
        # Create a new Layout instance for each repetition
        new_object = Layout(
            location=new_location,
            min=original.min,
            max=original.max,
            orientation=original.orientation
        )
        repeated_objects.append(new_object)
    
    return repeated_objects
    
def add_camera(location: Tuple[float, float, float], target_point: Tuple[float, float, float], lens: float = 35) -> bpy.types.Object:
    """
    Add a camera to the Blender scene.

    Args:
        location (Vector): The location to place the camera.
        target_point (Vector): The point the camera should be aimed at.
        lens (float, optional): The lens size. Defaults to 35.

    Returns:
        bpy.types.Object: The created camera object.

    Example:
        camera = add_camera((10, 10, 10), (0, 0, 0))
    """
    # Create a new camera data object
    cam_data = bpy.data.cameras.new(name="Camera")
    cam_data.lens = lens  # Set the lens property

    # Create a new camera object and link it to the scene
    cam_object = bpy.data.objects.new('Camera', cam_data)
    bpy.context.collection.objects.link(cam_object)

    # Set the camera location
    cam_object.location = location

    # Calculate the direction vector from the camera to the target point
    direction = Vector(target_point) - Vector(location)
    # Orient the camera to look at the target point
    rot_quat = direction.to_track_quat('-Z', 'Y')
    cam_object.rotation_euler = rot_quat.to_euler()

    # Set the created camera as the active camera in the scene
    bpy.context.scene.camera = cam_object

    return cam_object

\end{lstlisting}
\end{figure*}

\section{Examples of annotated queries}\label{sec:constrain}
Examples of the annotated queries as well as the per-scene scoring functions are shown in Figure~\ref{fig:human_1} and Figure~\ref{fig:human_2}.

\begin{figure*}[h!]
\begin{lstlisting}
scene_1 = {
    "description": "A book lying flat on a table, two chair on each side",
    "assets": ["book", "table"],
    "relationships": {
        "relativity": {
            "description": "The book should be on top of the table",
            "involved_objects": ["book", "table"]
        },
        "alignment": {
            "description": "The book should be aligned with the table in the x and y directions",
            "involved_objects": ["book", "table"]
        }
    }
}

def score_1(locs):
    # Extracting locations
    x_book, y_book, z_book = locs['book']['x'], locs['book']['y'], locs['book']['z']
    x_table, y_table, z_table = locs['table']['x'], locs['table']['y'], locs['table']['z']
    # Relativity score (penalizing if book is below table surface)
    relativity_score = max(0, z_table - z_book)  # positive if book is below table
    # Alignment score (difference in x and y positions, zero if perfectly aligned)
    alignment_score_x = abs(x_book - x_table)
    alignment_score_y = abs(y_book - y_table)
    # Total score (sum of individual scores)

    total_score = relativity_score + alignment_score_x + alignment_score_y
    return 1 -  total_score / 100.

scene_2 = {
    "description": "A busy airport terminal with people, seating areas, and information displays",
    "assets": ["person1", "person2", "seating_area", "information_display"],
    "relationships": {
        "grouping": {
            "description": "People should be grouped near the seating areas",
            "involved_objects": ["person1", "person2", "seating_area"]
        },
        "alignment": {
            "description": "Information displays should be aligned above the seating areas",
            "involved_objects": ["seating_area", "information_display"]
        },
        "proximity": {
            "description": "People should be close to information displays for visibility",
            "involved_objects": ["person1", "person2", "information_display"]
        }
    }
}
def score_2(locs):
    def distance(a, b):
        return math.sqrt((a['x'] - b['x'])**2 + (a['y'] - b['y'])**2 + (a['z'] - b['z'])**2)

    # Grouping score (distance of people from seating areas)
    grouping_score = sum(distance(locs[p], locs['seating_area']) for p in ['person1', 'person2'])
    # Alignment score (information display above seating areas)
    alignment_score = abs(locs['seating_area']['y'] - locs['information_display']['y'])
    # Proximity score (people close to information displays)
    proximity_score = sum(distance(locs[p], locs['information_display']) for p in ['person1', 'person2'])
    
    # Total score
    total_score = grouping_score + alignment_score + proximity_score
    return 1 - total_score / 100.

\end{lstlisting}
\caption{Example of annotated queries and scoring function} \label{fig:human_1}
\end{figure*}

\begin{figure*}[h!]
\begin{lstlisting}
scene_3 = {
    "description": "Three boxes of different sizes, stacked on top of each other",
    "assets": ["box1", "box2", "box3"],
    "relationships": {
        "hierarchy": {
            "description": "The boxes should be in descending order of size from bottom to top",
            "involved_objects": ["box1", "box2", "box3"]
        },
        "layering": {
            "description": "The boxes should be placed one above the other",
            "involved_objects": ["box1", "box2", "box3"]
        }
    },
}

def score_3(locs):
    # Extracting locations
    z_box1, z_box2, z_box3 = locs['box1']['z'], locs['box2']['z'], locs['box3']['z']
    w_box1, w_box2, w_box3 = locs['box1']['w'], locs['box2']['w'], locs['box3']['w']
    # Hierarchy score (sizes)
    hierarchy_score = 0
    if not (w_box1 > w_box2 > w_box3):
        hierarchy_score = abs(w_box1 - w_box2) + abs(w_box2 - w_box3)
    # Layering score (z-axis positioning)
    layering_score = 0
    if not (z_box1 < z_box2 < z_box3):
        layering_score = abs(z_box1 - z_box2) + abs(z_box2 - z_box3)
    # Total score
    total_score = hierarchy_score + layering_score
    return 1 - total_score / 100

scene_4 = {
    "description": "A new solar system with planets orbiting around a small star",
    "assets": ["sun", "planet1", "planet2", "planet3"],
    "relationships": {
        "rotation": {
            "description": "Planets should orbit around the sun",
            "involved_objects": ["planet1", "planet2", "planet3", "sun"]
        },
        "scaling": {
            "description": "Planets should vary in size",
            "involved_objects": ["planet1", "planet2", "planet3"]
        }
    }
}

def score_4(locs):
    import math
    def distance(a, b):
        return math.sqrt((a['x'] - b['x'])**2 + (a['y'] - b['y'])**2 + (a['z'] - b['z'])**2)

    # Rotation score (distance from sun)
    rotation_score = sum(distance(locs[p], locs['sun']) for p in ['planet1', 'planet2', 'planet3'])

    # Scaling score (size of planets)
    scaling_score = abs(locs['planet1']['size'] - locs['planet2']['size']) + abs(locs['planet2']['size'] - locs['planet3']['size'])

    # Total score
    total_score = rotation_score + scaling_score 
    return 1 - total_score / 100
\end{lstlisting}
\caption{Example of annotated queries and scoring function} \label{fig:human_2}
\end{figure*}

\section{Prompt Used at each stage}
The prompt used in SceneCraft is shown in Figure~\ref{fig:prompt}

\begin{figure*}[h!]
\begin{lstlisting}
query_find_assets = """I am writing several blender scripts to generate scenes for: %s. 
Please think step by step and then give me the assets (each is a single unit, avoid a composite set that contains multiple objects) that shall appear in these scenes. 
After explanation, structured in: Output: 1) x1: y1; 2) x2: y2; 3) ... Each with a general descriptive name (x) and a very detailed visual description (y)."""


query_height_assets = """I am writing several blender scripts to generate scenes for %s. 
Below are the assets we'd like to use. Now we need to scale them to correct height, please generate a python dictionary called height_dict, where key is each asset's name, and value is a number representing the height (measured in metre)
%s
Output the complete python dict via height_dict = {asset_name: height, ...}, also give detailed explanation as comment before the value in the dict. 
"""


query_plan_assets = """I am writing several blender scripts to generate a scene for %s.

Below are the assets I'd like to use:
%s

Now I want a concrete plan to put them into the scene. Please think step by step, and give me a multi-step plan to put assets into the scene. 
For each step, structure your output as:
layout_plan_i = {
    "title": title_i,
    "asset_list" : 
        [asset_name_1, asset_name_2], 
    "description": desc_i
}
where title_i is the high-level name for this step, and desc is detailed visual text description of what it shall look like after layout. asset_list is the non-empty list of assets to be added in this step.
Please think step by step, place assets from environmental ones to more details assets. Return me a list of python dictonaries layout_plan_1, layout_plan_2, ...
"""

prompt_graph = """
You are tasked with constructing a relational bipartite graph for a 3D scene based on the provided description and asset list. Your goal is to identify the spatial and contextual relationships between assets and represent these relationships in a structured format. Follow these steps:

1. Review the scene description and the list of assets.
2. Determine the spatial and contextual relationships needed to accurately represent the scene's layout. Consider relationships like proximity, alignment, parallelism, etc.
3. Construct the relational bipartite graph `G(s) = (A, R, E)` where:
   - `A` represents the set of assets.
   - `R` represents the set of relations as nodes.
   - `E` represents the edges connecting a relation node to a subset of assets `E(r)` in the scene that satisfies this relation.
4. For each identified relationship, create a relation node and link it to the appropriate assets through edges in the graph.

Output your findings in a structured format:
- List of relation nodes `R` with their types and descriptions.
- Edges `E` that link assets to their corresponding relation nodes.

This process will guide the arrangement of assets in the 3D scene, ensuring they are positioned, scaled, and oriented correctly according to the scene's requirements and the relationships between objects.
"""

\end{lstlisting}
\caption{Example of prompts being used}  \label{fig:prompt}\label{fig:human_2}
\end{figure*}

\begin{figure*}
    \centering
    \includegraphics[width=0.9\textwidth]{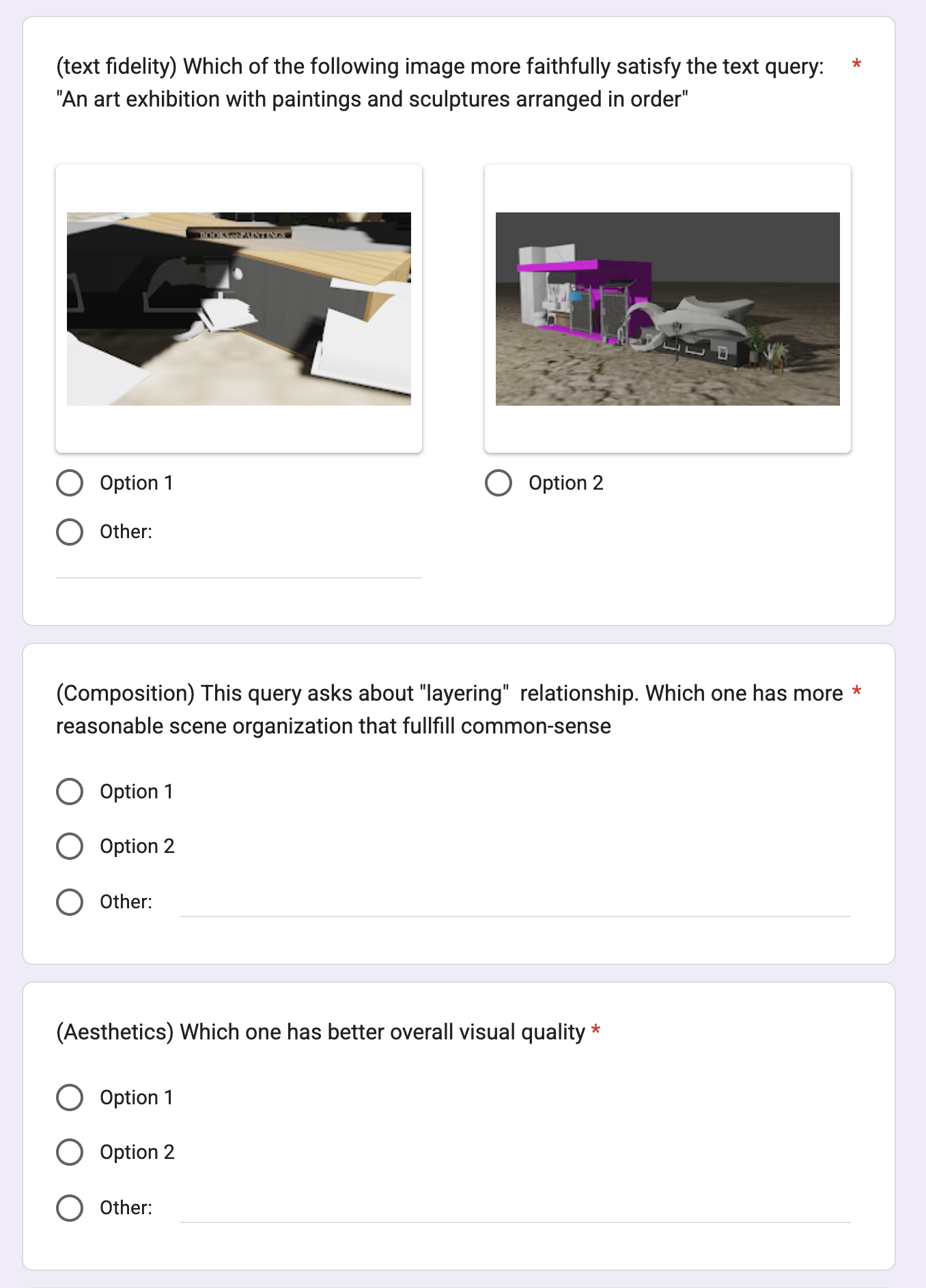}
    \caption{Questionnaire Interface, with three questions, about 1) Text Fidelity; 2) Composition; 3) Aesthetics}
    \label{fig:question}
\end{figure*}

\end{document}